%% file: sdmix-ubicomp.tex
\documentclass[acmlarge]{acmart}
\usepackage{algorithm}
\usepackage{algorithmic}
\usepackage{graphicx}
\usepackage{subfigure}
\usepackage{epstopdf}
\usepackage{xspace}
\usepackage{wrapfig}
\usepackage{algorithm}
\usepackage{algorithmic}
\usepackage{amsthm}

\usepackage{newfloat}
\usepackage{listings}
\usepackage{xspace}
\usepackage{bibentry}
\usepackage{multirow}
\usepackage{booktabs}
\usepackage{wrapfig}
\usepackage{tikz,xcolor}
\usepackage{scalerel}

\usepackage{hyperref}

\AtBeginDocument{%
  \providecommand\BibTeX{{%
    \normalfont B\kern-0.5em{\scshape i\kern-0.25em b}\kern-0.8em\TeX}}}

\setcopyright{acmlicensed}
\acmJournal{IMWUT}
\acmYear{2022} \acmVolume{6} \acmNumber{2} \acmArticle{65} \acmMonth{6} \acmPrice{}\acmDOI{10.1145/3534589}

\newcommand{\equationname}{Eq.}
\newtheorem{definition}{Definition}

\newcommand{\method}{SDMix\xspace}

\definecolor{lime}{HTML}{A6CE39}

\begin{document}

\title{Semantic-Discriminative Mixup for Generalizable Sensor-based Cross-domain Activity Recognition}

\author{Wang Lu}
\orcid{0000-0003-4035-0737}
\affiliation{%
  \institution{Beijing Key Lab. of Mobile Computing and Pervasive Devices, Inst. of Computing Tech., CAS}
  \city{Beijing}
  \country{China}
}
\email{luwang@ict.ac.cn}

\author{Jindong Wang}
\orcid{0000-0002-4833-0880}
\email{jindong.wang@microsoft.com}
\affiliation{%
  \institution{Microsoft Research Asia}
  \city{Beijing}
  \country{China}
}

\author{Yiqiang Chen}
\orcid{0000-0002-8407-0780}
\authornote{Wang Lu, Yiqiang Chen, and Xin Qin are also with University of Chinese Academy of Sciences. Yiqiang Chen is also with Pengcheng Laboratory. Correspondence to: Jindong Wang and Yiqiang Chen. }
\email{yqchen@ict.ac.cn}
\affiliation{%
  \institution{Beijing Key Lab. of Mobile Computing and Pervasive Devices, Inst. of Computing Tech., CAS}
  \country{China}
}

\author{Sinno Jialin Pan}
\orcid{0000-0001-6565-3836}
\email{sinnopan@ntu.edu.sg}
\affiliation{
    \institution{Nanyang Technological University}
    \country{Singapore}
}

\author{Chunyu Hu}
\orcid{0000-0002-3238-9888}
\email{hcy@qlu.edu.cn}
\affiliation{
    \institution{Qilu University of Technology (Shandong Academy of Sciences)}
    \country{China}
}

\author{Xin Qin}
\orcid{0000-0002-4717-4310}
\email{qinxin18b@ict.ac.cn}
\affiliation{%
  \institution{Beijing Key Lab. of Mobile Computing and Pervasive Devices, Inst. of Computing Tech., CAS}
  \city{Beijing}
  \country{China}
}
  
\renewcommand{\shortauthors}{Lu et al.}

\begin{abstract}

\end{abstract}

\begin{abstract}
It is expensive and time-consuming to collect sufficient labeled data to build human activity recognition (HAR) models. Training on existing data often makes the model biased towards the distribution of the training data, thus the model might perform terribly on test data with different distributions. Although existing efforts on transfer learning and domain adaptation try to solve the above problem, they still need access to unlabeled data on the target domain, which may not be possible in real scenarios. Few works pay attention to training a model that can generalize well to unseen target domains for HAR. In this paper, we propose a novel method called Semantic-Discriminative Mixup (SDMix) for generalizable cross-domain HAR. Firstly, we introduce semantic-aware Mixup that considers the activity semantic ranges to overcome the semantic inconsistency brought by domain differences. Secondly, we introduce the large margin loss to enhance the discrimination of Mixup to prevent misclassification brought by noisy virtual labels.
Comprehensive generalization experiments on five public datasets demonstrate that our \method substantially outperforms the state-of-the-art approaches with \textbf{6}\% average accuracy improvement on cross-person, cross-dataset, and cross-position HAR.
\end{abstract}



\maketitle

\section{Introduction}
\label{sec:Intro}

Sensor-based human activity recognition (HAR) aims to train machine learning models to recognize human activities based on different sensor data such as accelerometer and gyroscope.
HAR has wide applications in many areas including senior care, rehabilitation, and personal fitness~\cite{golestani2020human,wang2019deep}.
Machine learning HAR models, especially deep learning-based models, often need large amounts of well-labeled data for training.
However, it is expensive and time-consuming to collect or annotate massive labeled data in real applications.
Even if there exist sufficient data for training, the performance of models may still deteriorate when applied to a new environment with different data distributions.
For instance, a model trained on the activity data collected from the elderly cannot be directly used for adults because of the difference in their body size, activity patterns, and other characteristics.

\begin{figure}[htbp]
\centering
\subfigure[Different sensor readings on different subjects.]{
\label{fig-intro-peo}
\includegraphics[width=0.4\textwidth]{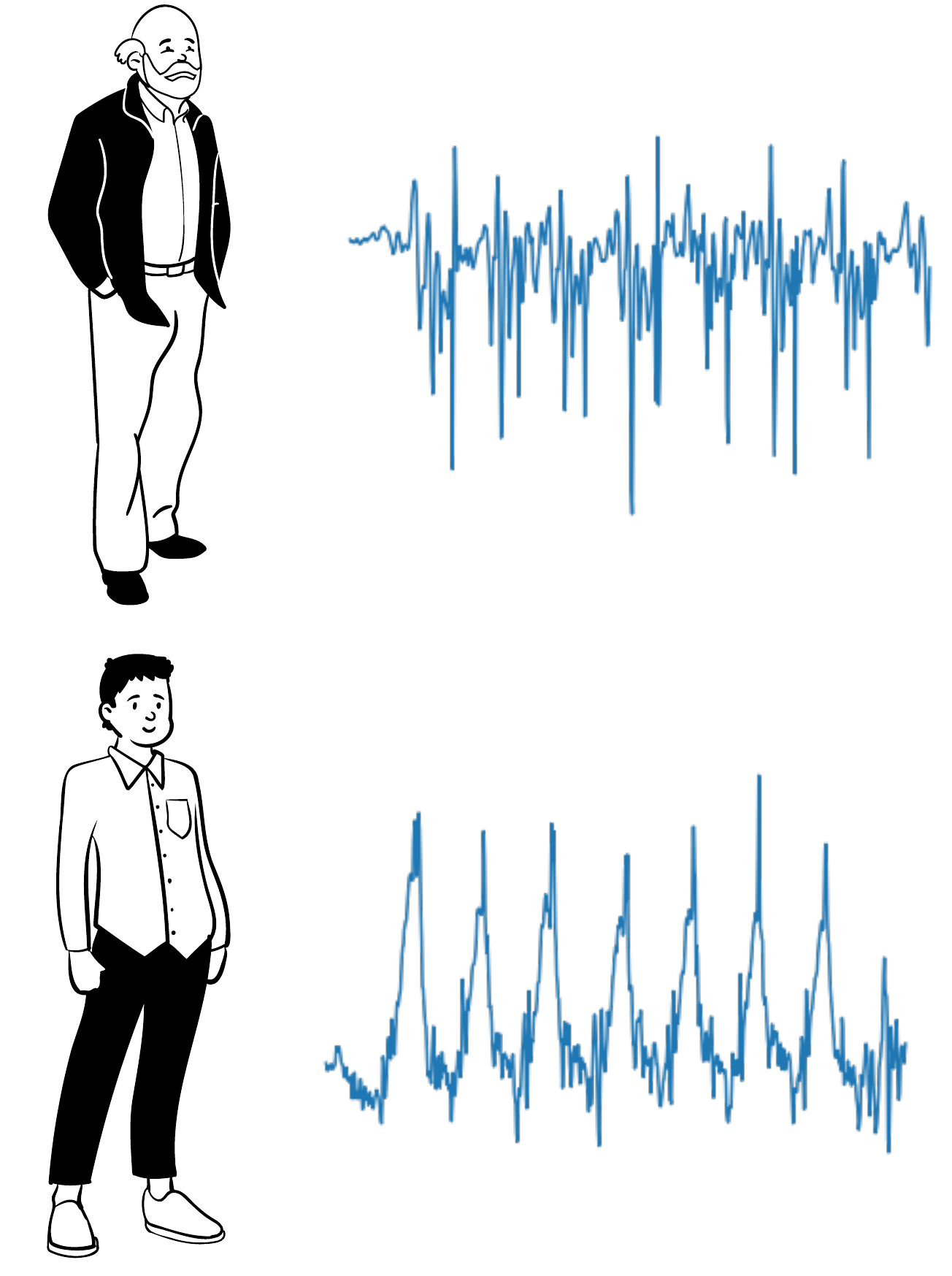}
}
\subfigure[Different sensor readings on different positions.]{
\label{fig-intro-pos}
\includegraphics[width=0.4\textwidth]{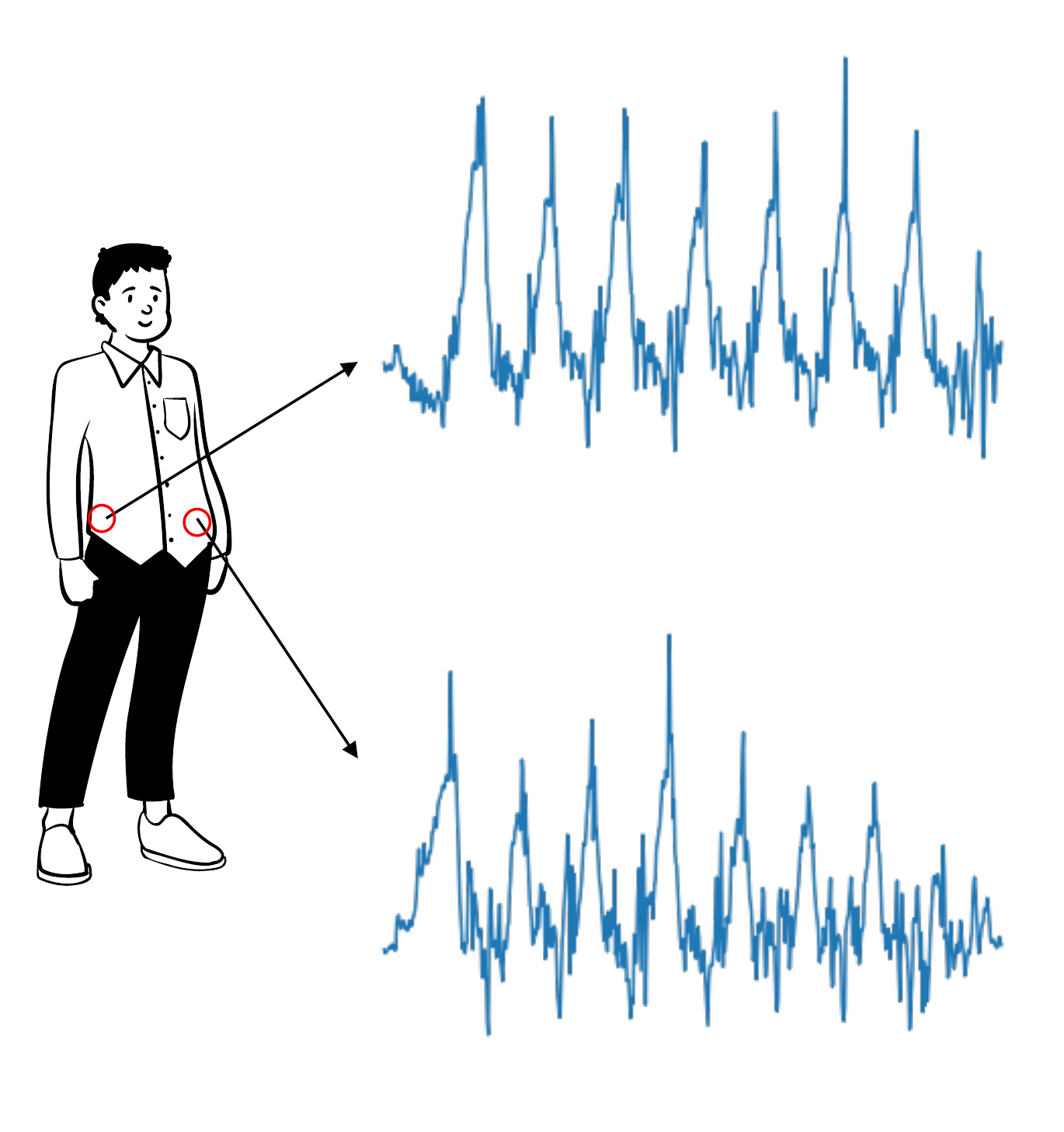}
}
\caption{Different sensor reading distributions on different subjects and different positions when walking.}
\label{fig:intro}
\end{figure}

As shown in \figurename~\ref{fig:intro}, the sensor data collected during walking follow different distributions. 
From \figurename~\ref{fig-intro-peo}, we can see data of the adult is more stable than data of the elderly when walking. 
And \figurename~\ref{fig-intro-pos} demonstrates that there exist differences between data collected from two positions of one person at the same time. 
Therefore, directly applying a model trained from existing data to a new environment may suffer from dramatic deterioration caused by distribution shifts.

Domain adaptation (DA)~\cite{wilson2020survey,pan2009survey} is a popular technique to transfer the knowledge from a well-labeled source domain to a target domain while reducing their distribution discrepancies.
Over the years, DA has been successfully applied to HAR by either aligning feature distributions via different distances~\cite{wang2018deep} or adversarial training~\cite{chang2020systematic}.
Nevertheless, DA needs access to the target data in training.
This may be less realistic in modern applications where we often expect a trained HAR model to perform well in different situations such as different environments, people, and datasets.
Recently, domain generalization (DG)~\cite{wang2021generalizing} is gaining increasing attention.
In contrast to DA, the goal of DG is to learn a model from one or several different but related domains that will generalize well on \emph{unseen} target domains.
For example, we expect a DG algorithm that trains on activity data from diverse ages to be able to generalize well to new persons of different ages.
Existing literature on DG is mainly based on data augmentation~\cite{wang2020heterogeneous}, domain-invariant learning~\cite{ganin2016domain}, or meta-learning~\cite{balaji2018metareg}.
Since HAR applications often require computation-restricted devices, we are interested in data augmentation-based DG methods for their simplicity and effectiveness.

Mixup~\cite{zhang2018mixup} is a popular data augmentation method and has been applied to DG~\cite{xu2021fourier,wang2020heterogeneous}.
Mixup enlarges the diversity of training data by interpolating between different domains.
However, the generalization performance of Mixup-based DG can be undermined because of two challenges: \emph{semantic inconsistency} and \emph{discriminative slackness}.
Firstly, the distributions of different activities are lying on different semantic spaces while direct Mixup will overlook their different semantic characteristics, resulting in semantic inconsistency between classes.
Secondly, the interpolation of mixup can easily generate noisy data, which will decrease discriminations of the activity classifier.

In this paper, we propose \textbf{\method}, a \emph{Semantic-Discriminative Mixup} approach for generalizable sensor-based HAR.
To prevent semantic inconsistency, \method utilizes a semantic-aware Mixup technique to prevent classification boundaries from approaching the activity categories with large distribution shifts.
We introduce activity semantic range for HAR that corresponds to the statistical values of activity shifts in the original space or feature space.
Therefore, this technique can eliminate negative effects brought by different semantic ranges of different activities.
Then, to handle discriminative slackness, \method introduces the large margin loss to replace the original cross-entropy loss.
Through the large margin loss, each class is far away from each other, which means most virtual data points will locate between classes and induce less noise.

To sum up, our contributions are as follows:

\begin{enumerate}
\item We explore Mixup for generalizable human activity recognition and empirically observe two limitations of Mixup: the semantic inconsistency and discriminative slackness, which undermine the performance of Mixup for domain generalization.
\item We propose \method to handle the semantic inconsistency and discriminative slackness challenges in generalizable HAR by introducing semantic-aware Mixup and enhancing its discrimination using large margin loss. \method is conceptually simple and easy to implement.
\item Extensive experiments demonstrate that our proposed \method significantly outperforms the state-of-the-art approaches (6\% average accuracy improvement on cross-person, cross-dataset, and cross-position scenarios).
\end{enumerate}

The remainder of this paper is organized as follows. Section~\ref{sec:relate} provides a brief review of the related work. Section~\ref{sec:formu} presents problem formulation and background. Then, Section~\ref{sec-method} proposes \method. In Section~\ref{sec:exp}, experimental evaluation and analysis on five public HAR datasets in three settings. Finally, we present the conclusions and the future work in Section~\ref{sec:concl}.

\section{Related Work}
\label{sec:relate}
\input{sec_related}

\section{Problem Formulation and Preliminaries}
\label{sec:formu}

We are given training data from several source domains. Our goal is to train a model that can perform well on an \emph{unseen} target domain with these available source domain data. 
Since the target data have different distributions compared to the source domains, we expect that the trained model is generalizable. 
A common technique to enhance generalization capability is data augmentation and Mixup is a common way. 
Mixup may encounter two challenges: semantic inconsistency and discriminative slackness.
To cope with these two challenges, SDMix is proposed which utilizes a semantic-aware Mixup technique to eliminate negative effects brought by different semantic ranges of different activities and introduces the large margin loss to reduce virtual noisy data.

\subsection{Problem Formulation}
Following the definition of generalizable cross-domain activity recognition from existing work~\cite{qian2021latent}, we are given
$S$ labeled source domains as the training dataset: $\mathcal{D}^{tr}=\{\mathcal{D}^i\}_{i=1}^S$.
We use $P^i(\mathbf{x},y)$ on $\mathcal{X} \times \mathcal{Y}$ to denote the joint distribution of one domain, where $\mathbf{x} \in \mathcal{X} \subset \mathbb{R}^{m}$ denotes the input and $y \in \mathcal{Y} = \{1, \cdots, C\}$ corresponds to output. $m$ and $C$ denote the input dimension and number of classes.
Our goal is to learn a generalized model from $\mathcal{D}^{tr}$ to predict well on an unlabeled target domain, $\mathcal{D}^T$, which is unseen in training.
In our problem, the training and test domains have the same input and output spaces but different distributions, i.e.,  $P^i(\mathbf{x},y) \neq P^j(\mathbf{x},y), \forall i,j \in \{1,2, \cdots, S, T\}$.
We aim to train a model $h$ from $\mathcal{D}^{tr}$ to minimize the risk on $\mathcal{D}^T$: $\min_h \mathbb{E}_{(\mathbf{x},y)\sim P^T} [h(\mathbf{x}) \neq y]$.

\subsection{Background}

Mixup~\cite{zhang2018mixup} is a popular approach for data augmentation.
Given two random samples $(\mathbf{x}_1, y_1), (\mathbf{x}_2, y_2)$, Mixup extends the training distribution by incorporating the prior knowledge: linear interpolations of feature vectors should lead to linear interpolations of the associated targets, formulated as \footnote{When performing Mixup, labels are often in the form of one-hot.}:
\begin{equation}
    \label{eqa:mix}
    \begin{aligned}
        \tilde{\mathbf{x}} & =\lambda \mathbf{x}_1 + (1-\lambda) \mathbf{x}_2,\\
        \tilde{y} & =  \lambda y_1 + (1-\lambda)y_2,
    \end{aligned}
\end{equation}
where $\lambda \sim Beta(\alpha,\alpha)$ and $\alpha \in (0, \infty) $ is a hyperparameter.
As a powerful data augmentation technique, Mixup plays a vital role to enlarge the distribution diversity in existing domain generalization research~\cite{wang2020heterogeneous,xu2021fourier}.


We are specifically interested in applying Mixup to generalizable HAR because of its simplicity.
Unfortunately in HAR, Mixup faces the semantic inconsistency and discriminative slackness challenges that will dramatically undermine its performance.

\section{\method}
\label{sec-method}

In this paper, we propose \textbf{\method} (Semantic-Discriminative Mixup) to solve these two challenges in generalizable HAR.
In the following, we will introduce the key components of \method.

\subsection{Semantic-aware Mixup}

In generalizable HAR, the distributions of sensor reading from two domains are different due to different lifestyles, habits, body shapes, and device positions.
While Mixup tries to increase the diversity of distributions, we argue that the \emph{activity semantic range} will largely impede Mixup.

\begin{definition}[Activity semantic range]
The activity semantic range of a domain $\mathcal{D}$ refers to the maximum shift towards the activity center, denoted as:
\begin{equation}
    R_c = \max d(\mu_{\mathbf{x} | y=c}, \mathbf{x}: \mathbf{x} \in P(\mathbf{x} | y = c)),
\end{equation}
where $\mu_{\mathbf{x} | y=c} = \mathbb{E}[\mathbf{x} | y=c]$ denotes the activity semantic center for class $c$ and $d(\cdot, \cdot)$ is a certain distance function.
\end{definition}

\begin{definition}[Semantic inconsistency]
If data of two classes have different activity semantic ranges, i.e., $R_{c_1}\neq R_{c_2}$, then we say that there exists a semantic inconsistency between these two classes.
\end{definition}

\begin{figure}[t!]
\centering
\vspace{-.1in}
\subfigure[Illustration (vanilla Mixup)]{
\label{fig-semantic-mixup}
\raisebox{0.35\height}{\includegraphics[width=0.4\textwidth]{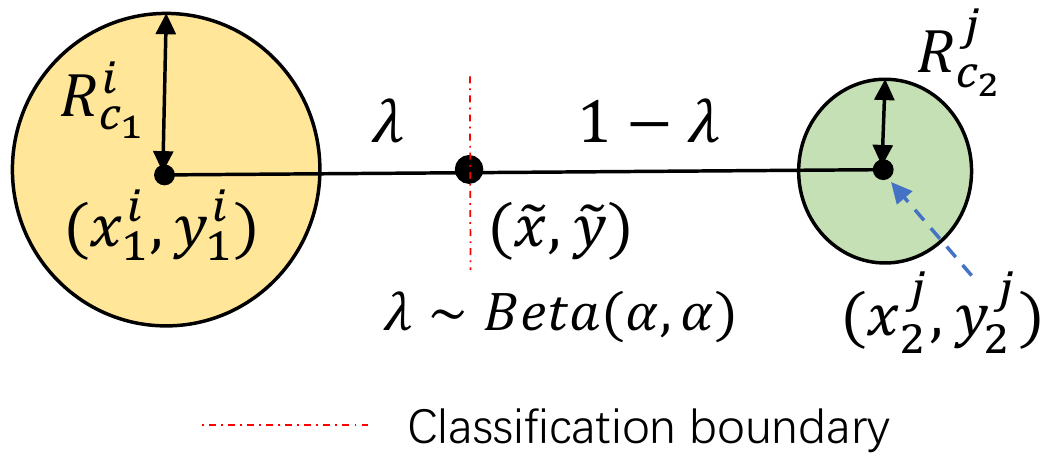}}

}
\subfigure[Decision boundary (vanilla Mixup)]{
\label{fig-dec-mixup}
\includegraphics[width=0.35\textwidth]{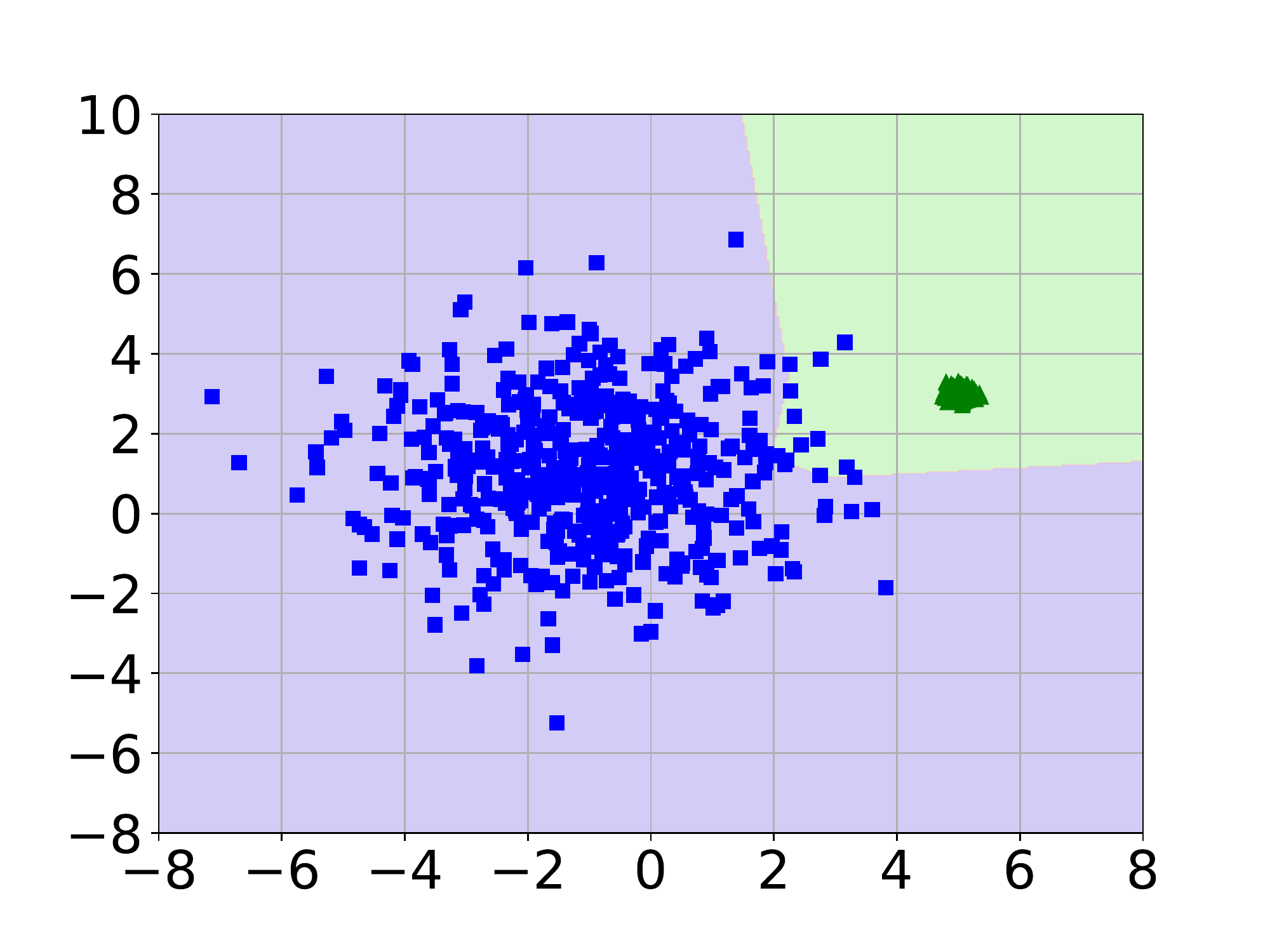}

}
\subfigure[Illustration (ours)]{
\label{fig-semantic-ours}
\raisebox{0.35\height}{\includegraphics[width=0.4\textwidth]{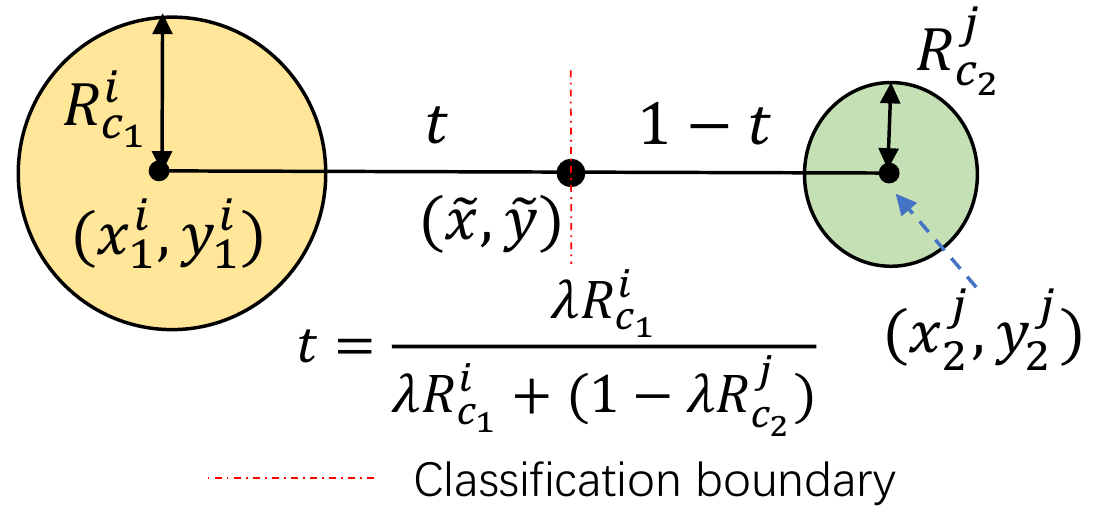}}
}
\subfigure[Decision boundary (ours)]{
\label{fig-dec-ours}
\includegraphics[width=0.35\textwidth]{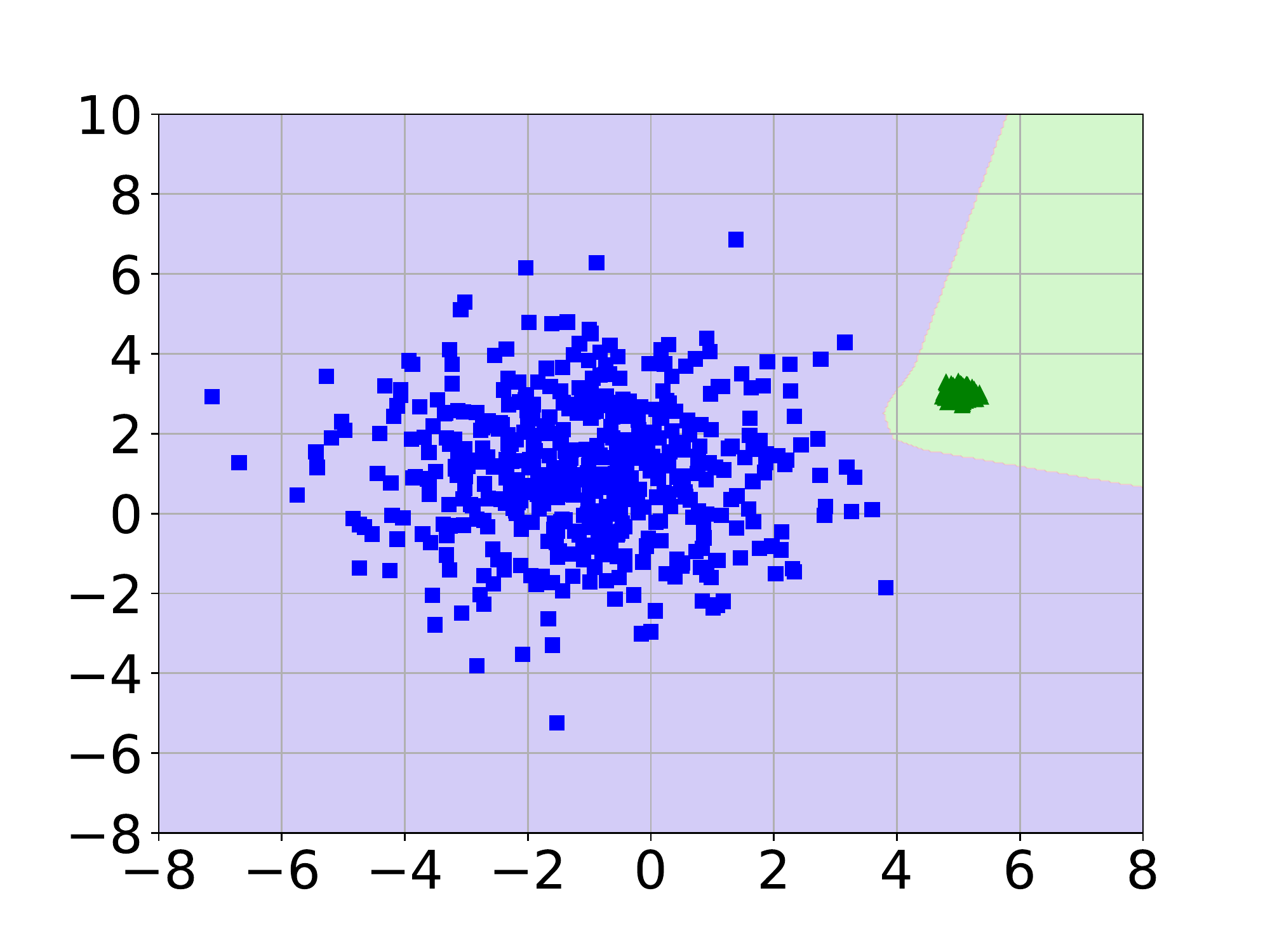}
}
\caption{The semantic inconsistency of Mixup. Different colors correspond to different classes. (a) Vanilla Mixup pays no attention to activity semantic range. (b) Decision boundary for vanilla Mixup leads to semantic inconsistency. (c) Our method considers the effects of activity semantic range. (d) Our semantic-aware Mixup mitigates such an issue.}
\label{fig:demo1}

\end{figure}

Different activities or even the same activity performed by different persons (domains) tend to have different activity semantic ranges.
For instance, a person that walks in a parking lot may have larger activity semantic range than another person that walks on a treadmill with a similar speed in flat since walking in a parking lot is more unstable and contains more degree of freedom. 
Another example is that two persons both playing basketball may have different activity semantic ranges due to different habits and positions.
The existing Mixup cannot consider the influence of activity semantic range from two different domains, resulting in \emph{semantic inconsistency}. 
Activity semantic range varies between any two classes from any different domains.
This situation cannot be avoided in both the feature space (\figurename~\ref{fig:vis}) and the original space.
As shown in \figurename~\ref{fig-semantic-mixup}, when the above situation occurs that two classes have different activity semantic ranges, the vanilla Mixup generates wrong classifications near the decision boundary.
Note that vanilla Mixup utilizes the same $\lambda$ for data and categories, which neglects the influence of activity semantic ranges. 

In this paper, we propose \emph{semantic-aware Mixup} to reduce the influence of different activity semantic ranges from different domains.
The samples generated by semantic-aware Mixup can be formulated as:
\begin{equation}
    \begin{aligned}
    &\tilde{\mathbf{x}}=\lambda \mathbf{x}^i_1 + (1-\lambda) \mathbf{x}^j_2, \\
    &\tilde{y} = t y^i_1 + (1-t)y^j_2,\\
        &\lambda \sim Beta(\alpha,\alpha),
    \end{aligned}
    \label{eqa:weightedmixup}
\end{equation}
where $(\mathbf{x}^i_1, y^i_1), (\mathbf{x}^j_2, y^j_2)$ denote samples from domain $i$ and $j$.
And $t$ is the \emph{activity semantic factor}, computed as:
\begin{equation}
    \label{eqa:getweight}
    t = \frac{\lambda \times R_{c_1}^i}{\lambda \times R_{c_1}^i + (1-\lambda) \times R_{c_2}^j},
\end{equation}
where $R_{c_1}^i$ and $R_{c_2}^j$ are the activity semantic ranges for the class $c_1$ and $c_2$ of domains $\mathcal{D}^i$ and $\mathcal{D}^j$, respectively. Note that $c_1$ and $c_2$ may or may not be the same, indicating that this works for both the same or different classes.
We only apply $t$ for $y$, which is equivalent to calibrating mixed $\tilde{y}$ with semantic range for a mixed $\mathbf{\tilde{x}}$.

What does Eq.~\eqref{eqa:getweight} do?
We illustrate its effects in \figurename~\ref{fig:demo1}, where the larger $R$ makes the corresponding $y$ play a more important role. 
In \figurename~\ref{fig-semantic-mixup}, vanilla Mixup pays no attention to the activity semantic range and treats $\mathbf{x}_i$ equally.
Therefore, its decision boundary is biased towards the class with larger activity semantic range.
While in \figurename~\ref{fig-semantic-ours}, the boundary is balanced by considering the effects of different activity semantic ranges. 
In \figurename~\ref{fig-dec-mixup} and \figurename~\ref{fig-dec-ours}, a toy example where simple models learned from simulation data following Gaussian distributions is to demonstrate that \method can learn better decision boundaries.

We present two alternatives to compute the value of activity semantic range $R_c^i$ by either using the largest distance from samples to centroid, or the mean of all distances:
\begin{subequations}
    \begin{align}
        R_c^i = \max_{\mathbf{x} \in \mathcal{D}^i_c} d(\mathbf{x}, \mu_{c}^i) \label{eqa:maxr},\\
        R_c^i = \frac{\sum_{\mathbf{x} \in \mathcal{D}^i_c} d(\mathbf{x}, \mu_{c}^i)}{|\mathcal{D}^i_c|}, \label{eqa:avgr}
    \end{align}
\end{subequations}
where $\mathcal{D}^i_c$ represents the samples associated with class $c$ from domain $\mathcal{D}^i$ and $|\cdot|$ is the cardinality.
For computation, $d(\cdot, \cdot)$ can be $\ell_1$, $\ell_2$-distances or cosine distance.
We choose different computing functions to achieve better performance.
$\mu_{c}^i$ denotes the $c$-th class center of domain $\mathcal{D}^i$:
\begin{equation}
    \mu_{c}^i=\frac{1}{|\mathcal{D}_c^i|} \sum_{\mathbf{x} \in \mathcal{D}_c^i}\mathbf{x}.
    \label{eqa:mu}
\end{equation}

\subsection{Enhancing Discrimination of Mixup}

Other than activity semantic range that describes the activity shift in HAR, we also notice another limitation of Mixup: the \emph{virtual noisy activity instance} will influence the discrimination of Mixup.

\begin{definition}[Virtual noisy activity instance]
The interpolation $(\tilde{\mathbf{x}},\tilde{y})$ generated by the Mixup of $(\mathbf{x}_1,y_1)$ and $(\mathbf{x}_2,y_2)$ is a virtual noisy activity instance, if
\begin{equation}
    ||h(\tilde{\mathbf{x}})-\tilde{y}||_1\geq \epsilon,
\end{equation}
where $h$ denotes the classification function and $\epsilon > 0$ is a threshold.
\end{definition}

\begin{definition}[Discriminative slackness]
When trained the model with virtual noisy activity instances via Vanilla Mixup, the model may fail to discriminate different classes. We say that the model meets the discriminative slackness problem.
\end{definition}

\begin{figure}[htbp]
\centering
\subfigure[Discrimination (vanilla Mixup)]{
\label{fig:demo3-1}
\includegraphics[width=0.38\textwidth]{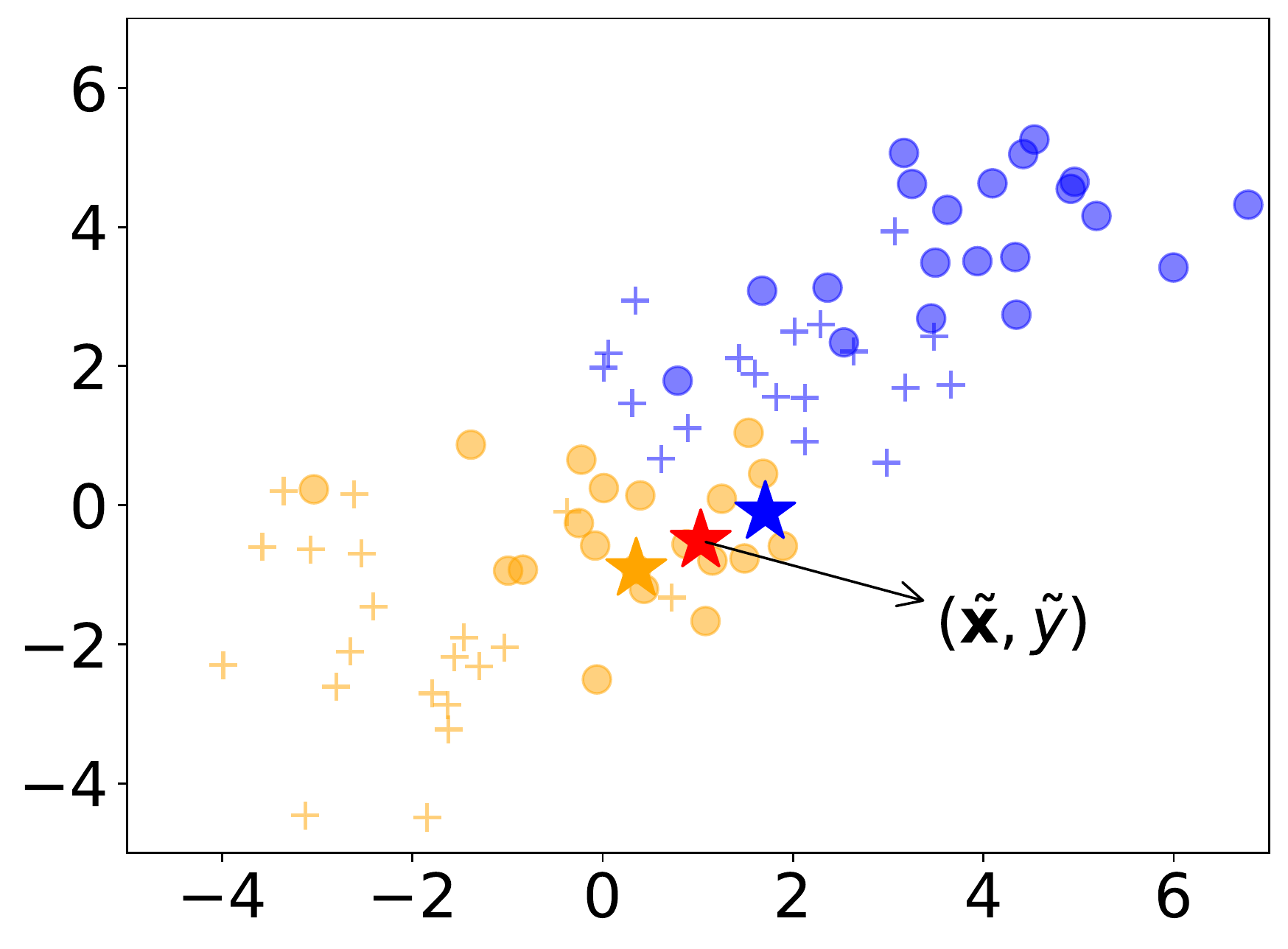}
}
\subfigure[Discrimination (ours)]{
\label{fig:demo3-2}
\includegraphics[width=0.38\textwidth]{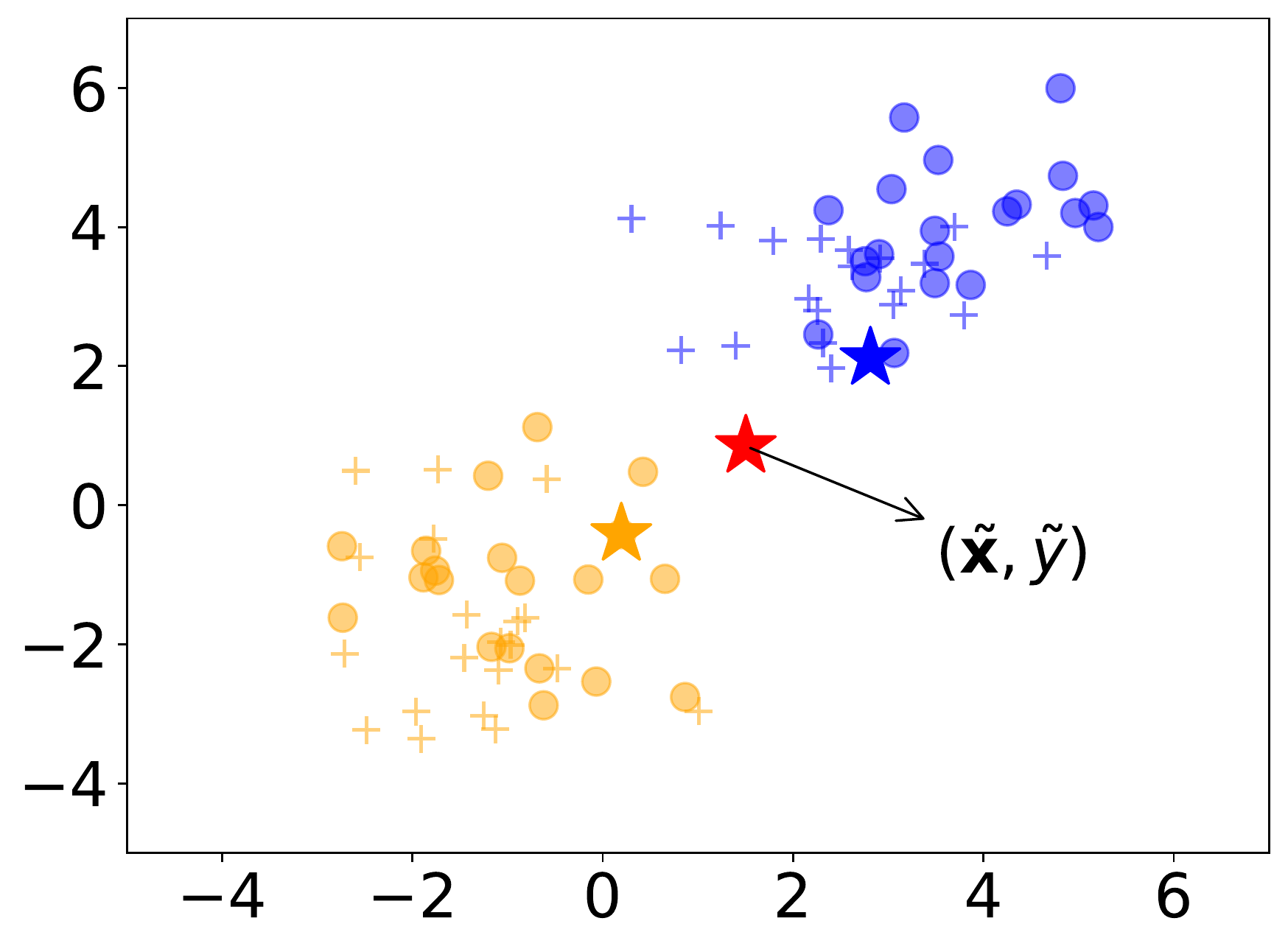}
}
\caption{Discriminative slackness of Mixup. The generated data in (a) can easily be misclassified. Different colors correspond to different classes, and different shapes correspond to different domains. Stars in blue or yellow are selected data points for mixing. The red star is the virtual data point. \emph{Best viewed in color.}}
\label{fig-illu-dis}
\end{figure}

Specially speaking, when $\mathbf{x}_1$ is close to the ideal decision boundary of class $y_2$ (or $\mathbf{x}_2$ to $y_1$), $\tilde{\mathbf{x}}$ could be extremely close to the decision boundary of $y_1$ or $y_2$ instead of their interpolation $\tilde{y}$.
And from the point of location, $\tilde{\mathbf{x}}$ may belong to $\hat{y}$ which is different from $y$.
For instance, due to different lifestyles, the running data of one person may have distributions similar to the walking data of another person.
Thus, his data is easily misclassified.
This causes \emph{discriminative slackness} of Mixup, as shown in \figurename~\ref{fig:demo3-1}.

In this paper, we introduce \emph{large margin loss} to enhance the discrimination of Mixup, which is formulated as:
\begin{equation}
    \ell_y(\mathbf{x}_k,y_k) = \mathscr{A}_{c \neq y_k} \max\{0,
    \gamma+ u_{h,\mathbf{x}_k,\{c,y_k\}} \mathrm{sign}(h_c(\mathbf{x}_k) - h_{y_k}(\mathbf{x}_k) \},
\label{eqa:margin}
\end{equation}
where $\ell_y$ is the large margin loss, $\mathscr{A}$ is an aggregation operator for multi-class setting, and $\mathrm{sign}(\cdot)$ adjusts the polarity of the distance. $h_c: \mathcal{X} \rightarrow \mathbb{R}$ or $h_{y_k}$ is a function that generates a prediction score for classifying the input vector $\mathbf{x}\in\mathcal{X}$ to class $c$ or $y_k$. $\gamma$ is the distance to the boundary which we expect. $u_{h,\mathbf{x},\{c_1,c_2\}}$
is the distance of a point $\mathbf{x}$ to the decision boundary of class $c_1$ and $c_2$, which is computed as:
\begin{equation}
    u_{h,\mathbf{x},\{c_1,c_2\}} = \min_{\delta} ||\delta||_p, \quad s.t.~ h_{c_1}(\mathbf{x}+\delta) = h_{c_2}(\mathbf{x}+\delta),
    \label{eqa:distbound}
\end{equation}
where $||\cdot||_p$ is $l_p$ norm. As shown in~\cite{elsayed2018large}, define $q = \frac{p}{p-1}$, then \equationname~\eqref{eqa:margin} can be computed as:
\begin{equation}
\begin{aligned}
    \ell_y(\mathbf{x}_k,y_k) =  \mathscr{A}_{c \neq y_{k}} \max \left\{ 0,  \gamma+\frac{h_{c}\left(\mathbf{x}_{k}\right)-h_{y_{k}}\left(\mathbf{x}_{k}\right)}{\left\|\nabla_{\mathbf{x}} h_{c}\left(\mathbf{x}_{k}\right)-\nabla_{\mathbf{x}} h_{y_{k}}\left(\mathbf{x}_{k}\right)\right\|_{q}} \right\}.
\end{aligned}
\label{eqa:approx}
\end{equation}

We illustrate the effect of large margin in \figurename~\ref{fig-illu-dis}. In \figurename~\ref{fig:demo3-1}, two classes with shape circle are far away from each other, but yellow circle data points are close to blue plus points.
Direct adoption of the vanilla Mixup with the blue star and the yellow star will generate the red star, which is a virtual noisy data point (according to the location of data points).
This means that a small class margin may induce terrible virtual labels when performing Mixup. In \figurename~\ref{fig:demo3-2}, we can see large margin can alleviate this issue.

\subsection{\method for HAR}

Similar to the vanilla Mixup, our proposed \method remains conceptually simple and can be easily trained in an end-to-end manner.
More importantly, our \method considers the semantics for different domains and also reduces noisy data when performing Mixup. 
Therefore, it makes the model perform better on unseen target data.

\begin{figure*}[htbp]
\centering
\includegraphics[width=0.9\textwidth]{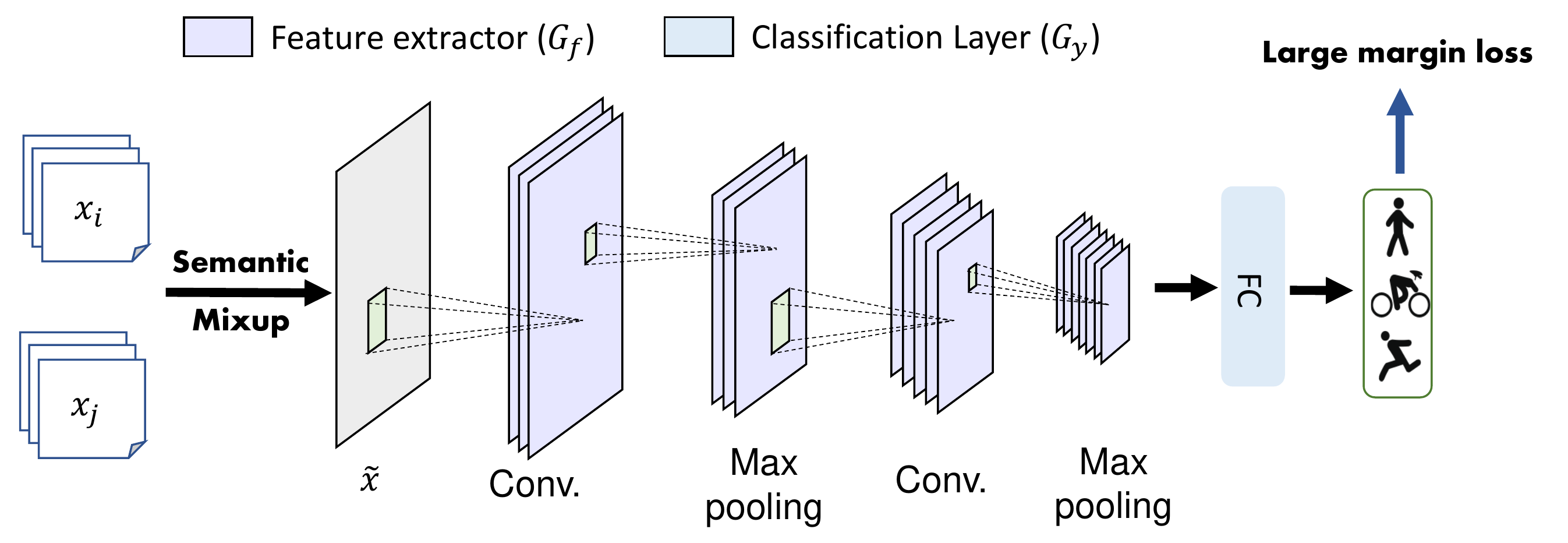}
\caption{The network architecture of the proposed method.}
\label{fig:frame}
\end{figure*}

\figurename~\ref{fig:frame} shows the network architecture of the proposed method.
Mixed data goes through two blocks and each block contains one convolution layer and one pooling layer acting as feature extractor $G_f$. 
We view 1D time series as 2D data with the height being 1, following Fig. 2. in~\cite{wang2019deep}.
Then, it goes through one fully-connected layer which serves as the classification layer $G_y$.
Semantics are fused via weighted label Mixup.
In \figurename~\ref{fig:frame}, we only show conducting Mixup in the original space.
We also can perform Mixup on the outputs of the feature extractor which can bring better results in some circumstances.
In addition,  we do not show BN layers in \figurename~\ref{fig:frame}.

The loss of \method can be expressed as:
\begin{equation}
    \ell_{\mathrm{SDMix}}(\mathbf{x}_1,y_1;\mathbf{x}_2,y_2)=\ell_y(\mathrm{Mix}_{\lambda}(\mathbf{x}_1,\mathbf{x}_2),\mathrm{Mix}_{\lambda}(y_1,y_2)),
\end{equation}
where $\mathrm{Mix}_\lambda(\cdot, \cdot)$ represents the semantic-aware mix operation computed by \equationname~\eqref{eqa:weightedmixup}. 
$\boldsymbol{z}$ is the outputs of the feature extractor. $\ell_y$ is the large margin loss.

Therefore, our learning objective is formulated as:
\begin{equation}
\min ~ \mathbb{E}_{P_1,P_2\sim P_\mathcal{D}} \mathbb{E}_{(\mathbf{x}_1,y_1)\sim P_1, (\mathbf{x}_2,y_2)\sim P_2 } \mathbb{E}_{\lambda \sim Beta(\alpha,\alpha)}\left[l_{\mathrm{SDMix}}(\mathbf{x}_1,y_1;\mathbf{x}_2,y_2)\right],
\end{equation}
where $P_\mathcal{D}$ is the distribution of domains and $P_1, P_2$ are distributions of data in selected domains.
Note that such an objective is optimized in a mini-batch style in deep networks and the details will be introduced in experiments. 
Specifically, we choose two source domains randomly each time and choose two samples from these two selected source domains respectively to perform semantic-aware Mixup. 
Semantic factors are updated every several iterations.
For fairness (the same number of optimization iterations), the model only utilizes interpolated examples and does not utilize the original training examples.
We believe that our method can achieve better performance with both original data and virtual data generated via mixing.

\section{Experiments}
\label{sec:exp}

We evaluate the proposed method by constructing different generalization scenarios on five public datasets.

\subsection{Datasets}

The statistical information of the five public datasets is shown in \tablename~\ref{tb-dataset}, which includes sensor types, sampling frequency, and activity classes.
Now, we briefly introduce them in the following.

The UniMib SHAR dataset (\textbf{SHAR})~\cite{micucci2017unimib} records 9 types of activities of daily living and 8 types of falls. 30 subjects (aged from 18 to 60) conduct the 17 fine-grained activities with data collected by an acceleration sensor embedded in an Android phone.
17 activities include 9 types of activities of daily living (StandingUpFL, LyingDownFS, StandingUpFS, Running, SittingDown, GoingDownS, GoingUpS, Walking, Null) and 8 types of falls (FallingBackSC, FallingBack, FallingWithPS, FallingForw, FallingLeft, FallingRight, HittingObstacle, Syncope).

UCI daily and sports dataset (\textbf{DSADS})~\cite{barshan2014recognizing} consists of 19 activities collected from 8 subjects wearing body-worn sensors on 5 body parts.
19 activities include sitting, standing, lying on back and on right side, ascending and descending stairs, standing in an elevator still, moving around in an elevator, walking in a parking lot, walking on a treadmill with a speed of 4 km/h, running on a treadmill with a speed of 8 km/h, exercising on a stepper, exercising on a cross trainer, cycling on an exercise bike in horizontal and vertical positions, rowing, jumping, and playing basketball.
Each subject wears three sensors: accelerometer, gyroscope, and magnetometer.

PAMAP2 physical activity monitoring dataset (\textbf{PAMAP2})~\cite{reiss2012introducing} contains data of 18 different physical activities, performed by 9 subjects wearing 3 sensors. 
18 activities include lying, sitting, standing, walking, running, cycling, Nordic walking, watching TV, computer work, car driving, ascending stairs, descending stairs, vacuum cleaning, ironing, folding laundry, house cleaning, playing soccer, rope jumping, and other (transient activities).
The sampling frequency is 100Hz and the data dimension is 27 (3 inertial measurement units).

USC-SIPI human activity dataset (\textbf{USC-HAD})~\cite{zhang2012usc} is composed of 14 subjects (7 male, 7 female, aged from 21 to 49) executing 12 activities with a sensor tied on the front right hip. 
The data dimension is 6 and the sample rate is 100Hz.
12 activities include Walking Forward, Walking Left, Walking Right, Walking Upstairs, Walking Downstairs, Running Forward, Jumping Up, Sitting, Standing, Sleeping, Elevator Up, and Elevator Down.

Human activity recognition using smartphones dataset (\textbf{UCI-HAR})~\cite{anguita2012human} is collected by 30 subjects performing 6 daily living activities with a waist-mounted smartphone.
6 activities include walking, sitting, lying, standing, walking upstairs, walking downstairs.
The data dimension is 6 and the sample rate is 50Hz.

\begin{table}[htbp]
\caption{Statistical information of five datasets.}
\label{tb-dataset}
\resizebox{.45\textwidth}{!}{%
\begin{tabular}{crrrr}
\toprule
Dataset & \#Subject & \#Sensor & \#Class & \#Sample  \\ \midrule
SHAR    & 30        & 1        & 17         & 236,919  \\
DSADS   & 8         & 3        & 19         & 1,140,000 \\
UCI-HAR     & 30        & 6        & 6         & 1,310,000 \\
PAMAP2   & 9         & 3        & 18         & 3,850,505 \\
USC-HAD     & 14        & 2        & 12         & 5,441,000 \\
\bottomrule
\end{tabular}%
}
\end{table}

\subsection{Experimental Setup}

We construct three types of experiments to thoroughly evaluate all methods for generalizable HAR: (1) Cross-person HAR similar to existing research~\cite{qian2021latent}, (2) Cross-dataset HAR, which is our construction that is more challenging than cross-person HAR since cross-dataset setting implies that not only the persons are different, but also the sensor types and positions, and (3) Cross-position HAR, which utilizes data from different positions with same persons.
In all three settings, we adopt the sliding window technique~\cite{bulling2014tutorial} with 50\% overlap to construct training samples following common practice in HAR~\cite{wang2019deep}.
The inputs should be cut into individual inputs according to the sampling rate and each new input serve as one instance.

We give a more detailed introduction of three settings in the following.
\begin{itemize}
    \item \textbf{Cross-Person}: We conduct experiments on SHAR, DSADS, PAMAP2, and USC-HAD for this setting.
    We divide each dataset into 4 domains, e.g., 8 people are evenly divided into 4 domains where each domain consists of data from 2 people and there is no overlap between domains.
    For SHAR, we follow GILE to choose four persons. For the other datasets, we divide all persons into 4 groups.
    For PAMAP2, we choose 12 classes to ensure that each class has a certain number of data.
    \item \textbf{Cross-Dataset}: We conduct experiments on DSADS, PAMAP2, USC-HAD, and UCI-HAR for this setting.
    We do not split domains; instead, each dataset can be treated as one domain and we take the common classes from all datasets.
    To perform the Cross-Dataset setting, we first need to unify $\mathcal{X}$ and $\mathcal{Y}$.
    For $\mathcal{X}$, two sensors that belong to the almost same position are selected and each sample contains six dimensions.
    Then, data is down-sampled and the sampling rate is 25Hz.
    For $\mathcal{Y}$, six common classes are selected, including walking, walking upstairs, walking downstairs, sitting, standing, and lying.
    \item \textbf{Cross-Position}: We conduct experiments on DSADS for this setting.
    3 sensors are located in five parts of a body in DSADS.
    Each body part corresponds to a different domain.
    Each sample contains three sensors with nine dimensions.
\end{itemize}

\begin{table}[htbp]
\centering
\caption{Detailed information on cross-person, cross-dataset, and cross-position experiments.}
\resizebox{0.8\textwidth}{!}{%
\begin{tabular}{cccccc}
\toprule
\multicolumn{6}{c}{Cross-Person}\\
\midrule
Dataset & \#Domain & \#Sensor & \#Class & \#Domain Sample& \#Total\\
\midrule
SHAR&4&1&17&(57,984;88,033;45,904;44,998)&236,919\\
DSADS&4&3&19&(285,000)$\times$4&1,140,000\\
PAMAP2&4&3&12&(592,600; 622,200; 620,000; 623,400)&2,458,200\\
USC-HAD&4&2&12&(1,401,400;1,478,000;1,522,800;1,038,800)&5,441,000\\
\midrule
\multicolumn{6}{c}{Cross-Dataset}\\
\midrule
Dataset & \#Domain & \#Sensor & \#Class & \#Domain Sample& \#Total\\
\midrule
& 4&2&6&(672,000;810,550;514,950;470,850)&2,468,350\\
\midrule
\multicolumn{6}{c}{Cross-Position}\\
\midrule
Dataset & \#Domain & \#Sensor & \#Class & \#Domain Sample& \#Total\\
\midrule
DSADS& 5&3&19&(1,140,000)*5&5,700,000\\
\bottomrule
\end{tabular}%
}
\label{tb-data-crossp-crossd}
\end{table}

\tablename~\ref{tb-data-crossp-crossd} shows the information on cross-person, cross-dataset and cross-position scenarios.
It is worth noting that since SHAR dataset contains much fewer samples than others, we replace it with UCI-HAR dataset~\cite{anguita2012human} for cross-dataset experiments. 
The reason we did not run cross-person experiments on UCI-HAR is that the results on this dataset are all near 100\%.
In total, we constructed 16, 4, and 5 tasks for cross-person, cross-dataset, and cross-position HAR, respectively.
We use $0, 1, 2, 3$ to denote the four divided domains for cross-person experiments.
We use $0, 1, 2, 3, 4$ to denote the five divided domains for cross-position experiments.

\paragraph{Baselines.}
The latest method for generalizable HAR is GILE (Generalizable Independent Latent Excitation)~\cite{qian2021latent}.
Since there are no other methods designed for HAR except GILE, we turn to comparing with several popular domain generalization methods.
DeepAll combines all source data together and trains a model according to ERM.
DANN~\cite{ganin2016domain} and Coral~\cite{sun2016deep} are two traditional DG methods, and we extend them to fit the domain generalization setting.
Mixup~\cite{zhang2018mixup}, GroupDRO~\cite{sagawa2019distributionally},
RSC~\cite{huang2020self}, and ANDMask~\cite{parascandolo2020learning} are four state-of-the-art universal domain generalization methods which can be directly used for HAR. Except GILE, we reproduced all other methods with the same network architecture in Pytorch~\cite{paszke2019pytorch} for fairness. 
GILE is based on VAE, and we directly use their released code.
\tablename~\ref{tab:my-table-kernelsize} shows information on the input size and kernel size of the models in all settings and the final dimension of input size represents the window size.
We do not compare other Mixup-based DG methods since they are built for computer vision tasks that are not applicable to HAR problems.

\begin{table}[htbp]
\centering
\caption{Information on the architectures of the models.}
\resizebox{0.45\textwidth}{!}{%
\begin{tabular}{cllc}
\toprule
Setting&Dataset&Input&Kernel Size\\
\midrule
 \multirow{4}{*}{Cross-Person} 
 &SHAR&(3,1,151)&(1,9)\\ 
 & DSADS&(45,1,125)   & (1,9) \\
 & PAMAP2&(27,1,200)&(1,9)\\
 &USC-HAD&(6,1,200)&(1,6)\\
Cross-position &-&(9,1,125)& (1,9) \\
Cross-dataset  &-&(6,1,50)& (1,6) \\
\bottomrule
\end{tabular}}
\vspace{-.1in}
\label{tab:my-table-kernelsize}
\end{table}

To obtain the final model, we split data of the source domains into the training splits and validation splits. 
The training splits are utilized for training the model while the validation splits are utilized to guide the selection of the model.
For all benchmarks, we select the best model via validation accuracy.
In practice, we leave 80\% of source domain data as training splits while the rest data are for validation.
For testing, we evaluate the selected models on all data of the target domain.
For our architecture, the model contains two blocks, and each has one convolution layer, one pool layer, and one batch normalization layer. 
A single fully-connected layer serves as the classifier. In each step, each domain selects 32 samples. The maximum training epoch is set to 150. 
For our method, we search best hyperparameters within $[0.1,0.2,0.5,1,10]$ for $\alpha$, $[1,2,5]$ for top $c$ in large margin loss, and $[10,100,10000,100000]$ for $\gamma$.
For all methods except GILE\footnote{The source code for GILE is available at \url{https://github.com/Hangwei12358/cross-person-HAR}}, the Adam optimizer with a learning rate $10^{-2}$ and weight decay $5\times 10^{-4}$ is used. 
We use outputs of the feature extractor to compute radii and utilize different computing ways for different datasets. 
For fair study, we extensively tune hyperparameters for each method and report their average results of three trials.
Code for \method will be available at \url{http://transferlearning.xyz}.

\subsection{Experimental Results}

\begin{figure*}[htbp]
\centering
\subfigure[Cross-person: nearly balanced classification]{
\label{fig:f1score}
\includegraphics[width=.48\textwidth]{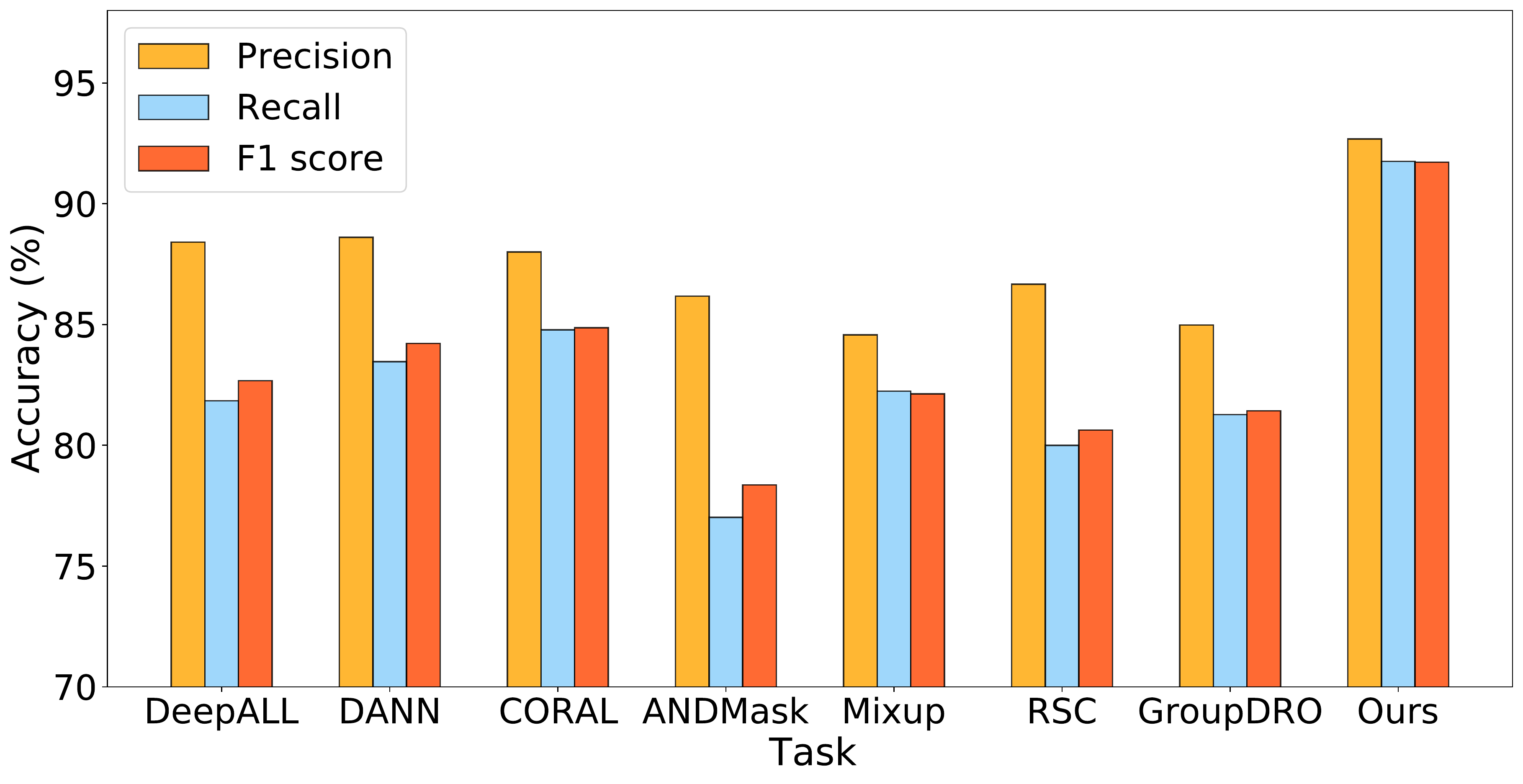}
}
\subfigure[Cross-dataset: imbalanced classification]{
\label{fig:f1score1}
\includegraphics[width=.48\textwidth]{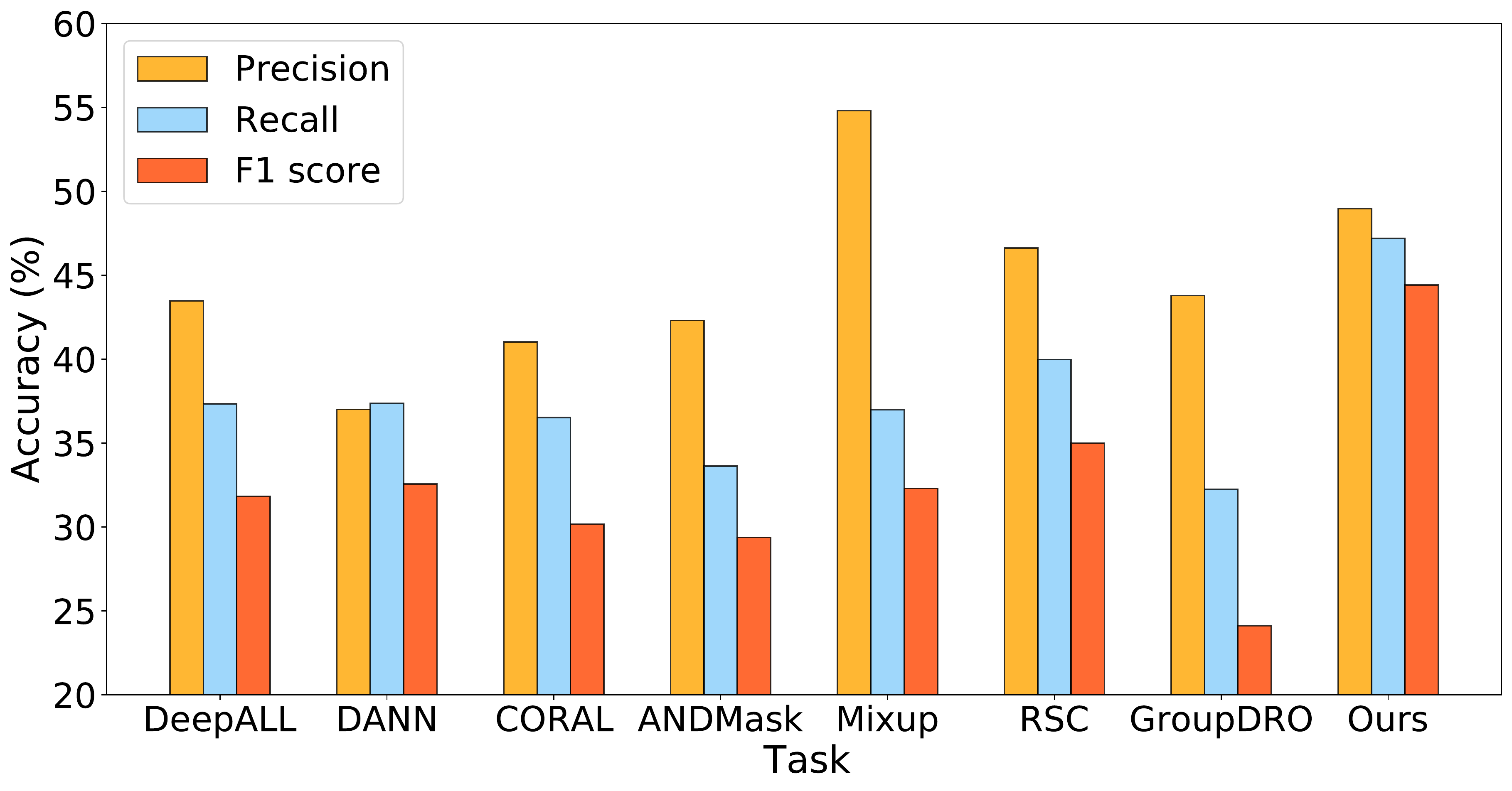}
}
\caption{Precision, recall, and F1 score.\figurename~\ref{fig:f1score} contains results from the first task in DSADS where data of the first two persons are used for testing and data of the rest persons are used for training. Each class has an equal number of instances (120). In \figurename~\ref{fig:f1score1}, data in USC-HAD are used for testing while data in other datasets are used for training. Classes in USC-HAD have 2545, 2290, 3608, 2048, 1904, 3744 instances respectively.}
\label{fig:f1}
\end{figure*}

The results of cross-person, cross-dataset, and cross-position HAR are shown in \tablename~\ref{tb-crossperson}, \ref{tb-crossdataset}, and \ref{tb-crossposition} respectively.
On average, our proposed \method substantially outperforms the second-best methods: \textbf{6.0\%} for cross-person, \textbf{10.33\%} for cross-dataset HAR, and \textbf{2.34\%} for cross-position HAR.
Moreover, from \figurename~\ref{fig:f1}, we can see that our method also achieves the best F1 score compared with other state-of-the-art methods whether in a balanced or unbalanced situation.
These indicate that our method is effective for generalizable cross-domain HAR applications.

\input{tb_allresults}
\input{tb_crossdataset}
\input{tb_crossposition}

We observe more insightful conclusions.
(1) \emph{When are Mixup-based methods most effective?}
When the dataset volume is rather small with more classes, i.e., on difficult tasks. This can be verified on SHAR and DSADS datasets that have fewer samples but more classes.
In these datasets, Mixup-based methods dramatically increase the diversity of data distributions that benefits generalization.
On the larger PAMAP2 and USC-HAD datasets that involve more diversity, the vanilla Mixup only slightly surpasses DeepAll or even worse.
(2) \emph{When is \method less effective?}
When the training domains are much larger than the test domains, i.e., the training domains are already containing diverse\footnote{Diversity is relative. Few instances may be enough for simple situations (e.g. iid) while it may need more instances for difficult cases.} data, where \method may bring smaller improvements.
In our statistics, source domains $1\sim3$ in PAMAP2 have more samples that make it enough to learn models directly using DeepAll. 
The same goes for cross-dataset HAR when UCIHAR is the target.
In addition, when activity semantic range is hard to compute, \method only slightly surpasses other methods or even worse.
For example, there exist little information and many categories in the cross-position setting, which makes it hard to estimate activity semantic range.
\method only has a slight improvement compared with other methods and even performs worse on the first task.
(3) \emph{What influences stability in HAR generalization?}
Algorithms, random states, data splits, and so on.
Whichever algorithm is used, there exist fluctuations among the results of three trials. Even with DeepAll, results are also different. 
From \tablename~\ref{tb-crossperson}, we can see that our method achieves the smallest STD compared to other state-of-the-art methods except GILE\footnote{GILE utilizes a different network architecture from other methods. Moreover, GILE selects the best model according to the target, which is unrealistic in reality and may be contrary to the principle of domain generalization.}.
The results demonstrate that our method is still reliable and stable with three trials compared to other state-of-the-art methods.
(4) \emph{What influences generalization in HAR?}
Dataset quantity, diversity, and the cross-domain distribution discrepancy.
There is no doubt that the generalization ability of a model will increase when the dataset becomes larger and diverse (rf. \tablename~\ref{tb-crossperson}).
More importantly, we see a significant performance drop from cross-person to cross-position HAR, indicating the importance of cross-domain discrepancies~\cite{chang2020systematic}.
The cross-dataset and position scenario is more challenging than cross-person since the sensor devices, positions, subjects are all different.
In this scenario, our \method achieves the best performance by harnessing their diversity.
Note that there is still room for improvement in this setting.

\begin{figure}[htbp]
\centering
\subfigure[Vanilla Mixup]{
\label{fig:confmixup}
\includegraphics[height=.25\textwidth]{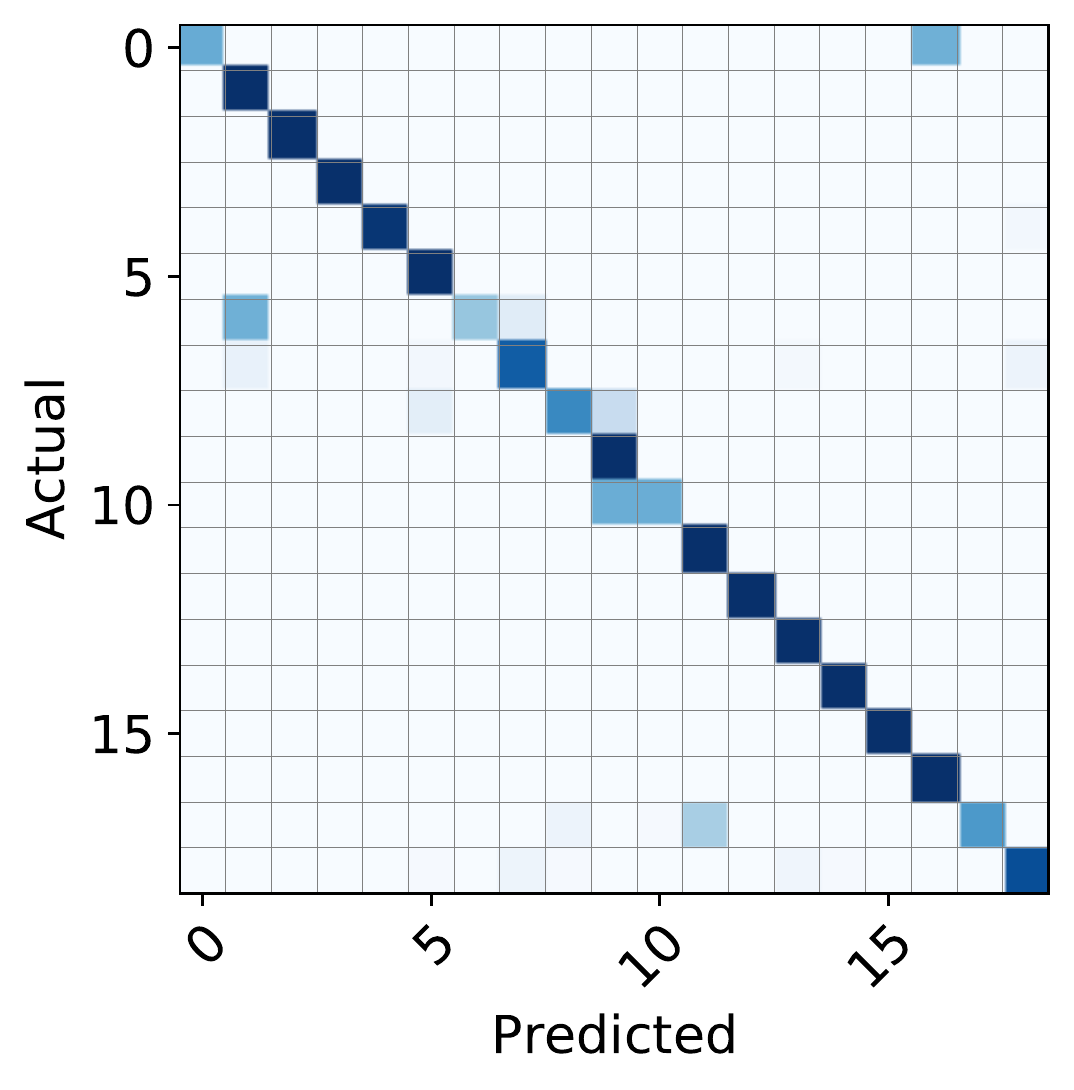}
}
\subfigure[Our method]{
\label{fig:confour}
\includegraphics[height=.25\textwidth]{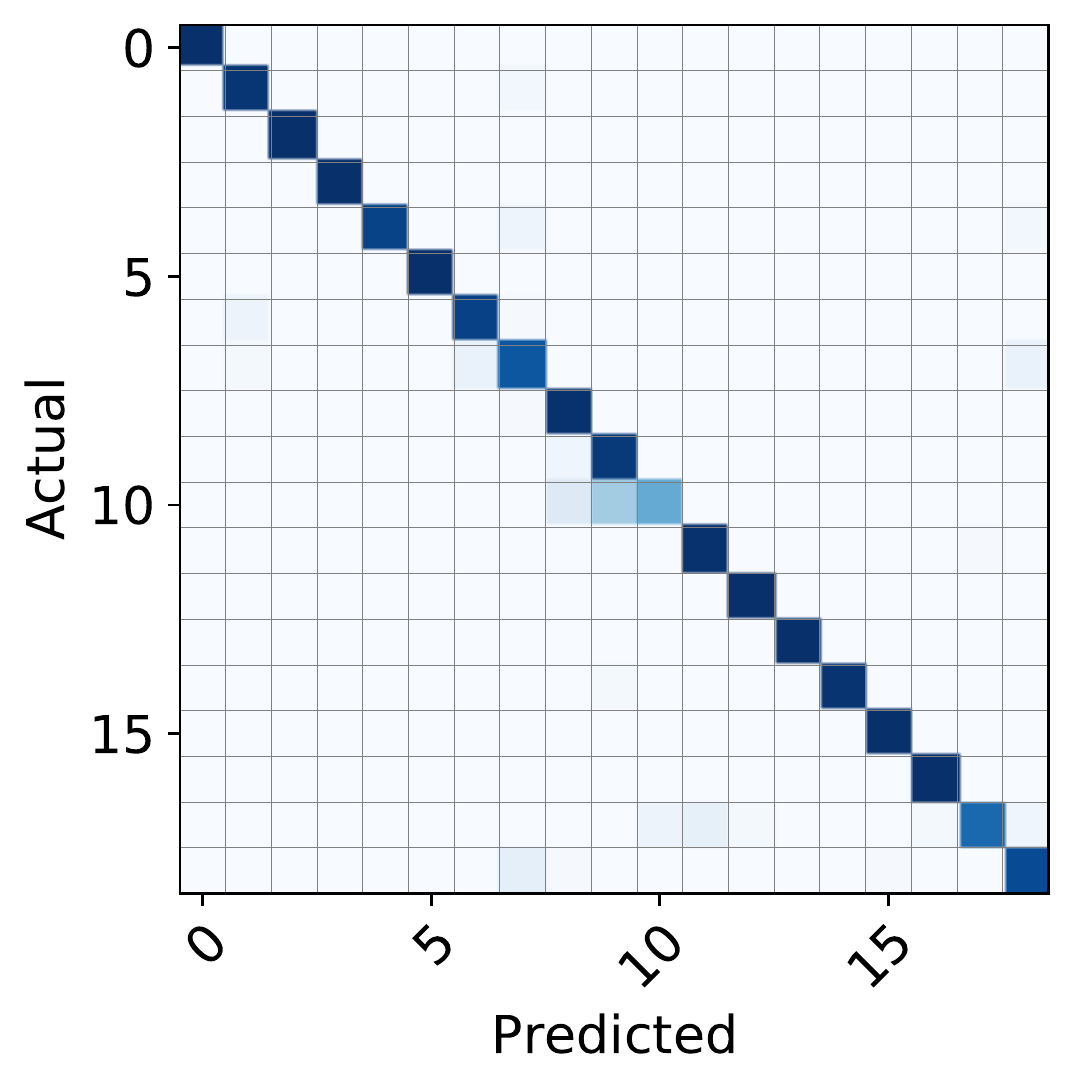}
}
\subfigure[Visualization]{
\label{fig:visd0}
\includegraphics[height=0.25\columnwidth]{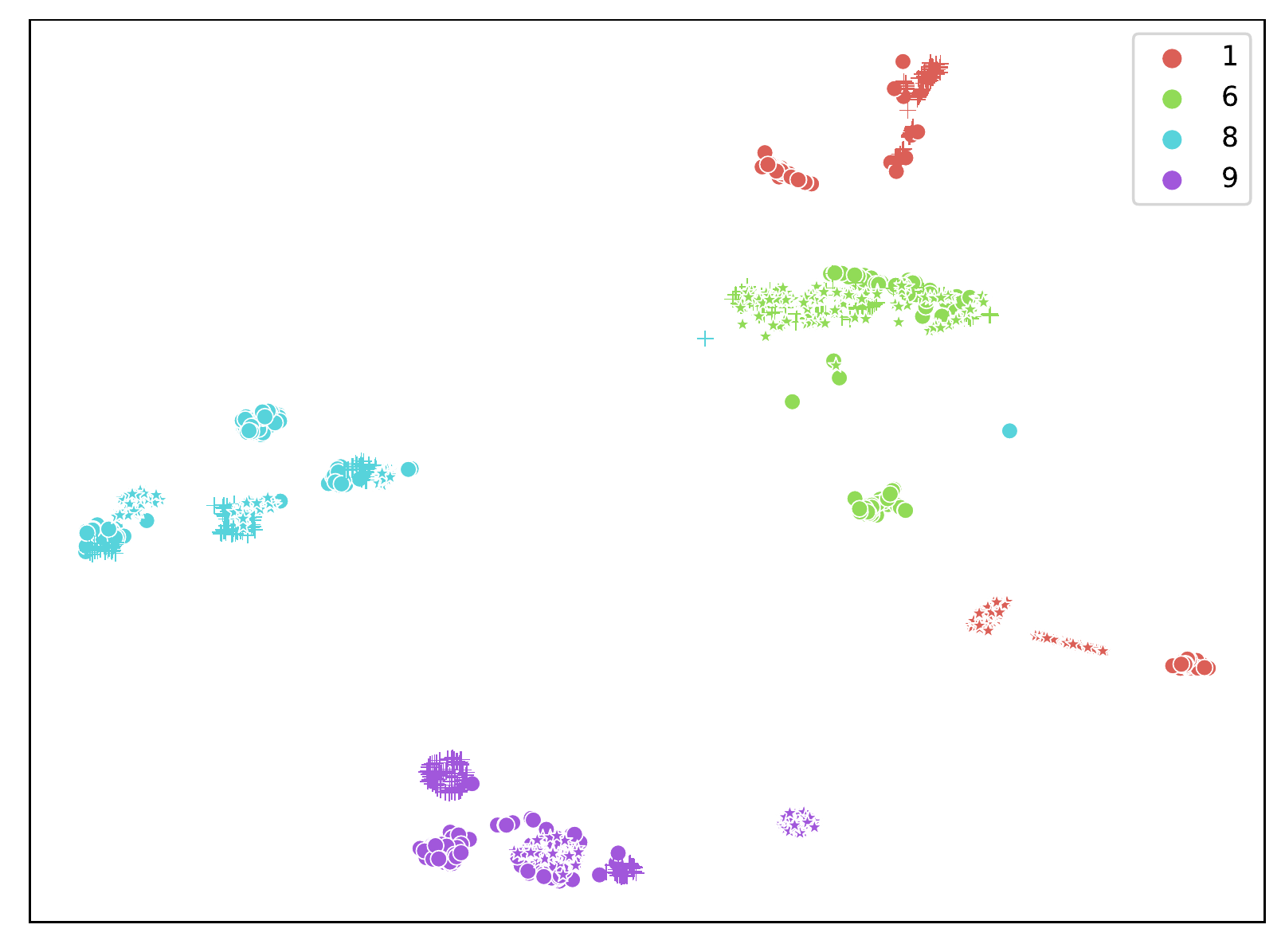}
}
\caption{Confusion matrices and Visualization on DSADS dataset for target 0 in the Cross-Person setting. \figurename~\ref{fig:confmixup} and \figurename~\ref{fig:confour} are confusion matrices of vanilla Mixup and our method. Deeper colors denote larger values. \figurename~\ref{fig:visd0} is the visualization of the t-SNE embeddings with features learned by a simple model. Each class is denoted by a color and each shape corresponds to a domain. \emph{Best viewed in color and zoom in.}}
\label{fig:conf}
\end{figure}

\subsection{Analysis}

Why is \method effective?
Do semantic inconsistency and discriminative slackness really exist?
To analyze the rationale behind the effectiveness of \method, we plot the confusion matrices for the vanilla Mixup and ours for target 0 on DSADS dataset in \figurename~\ref{fig:conf}.
While most classes are correctly classified, we are more interested in the misclassified classes: class 8 (walking in a parking lot) and class 9 (walking on a treadmill with a speed of 4km/h in flat); class 1 (standing) and class 6 (standing in an elevator still).
Firstly, as shown in \figurename~\ref{fig:confmixup}, some samples in class 8 are misclassified as class 9 (Please see the row labeled 8).
It is caused by different activity semantic ranges since walking in a parking lot has more variance than walking on a treadmill.
Our method can correctly classify them.
Secondly, some samples in class 6 are misclassified as class 1 (Please see the row labeled 6) since standing and standing in an elevator still are two similar activities with small differences that weaken the discrimination of Mixup. 
Again, our method can mitigate these two issues to achieve the best results as shown in \figurename~\ref{fig:confour}.
As shown in \figurename~\ref{fig:visd0}, class 8 in color 8 with shape circle has a larger semantic range than class 9 while class 1 almost compasses class 6 which illustrates they are close to each other.
\figurename~\ref{fig:visd0} describes the phenomena while confusion matrices are the results brought by the phenomena, indicating the claim mentioned above.

\begin{figure}[b!]
\centering
\subfigure[Semantic Range]{
\label{fig:vism}
\includegraphics[width=0.75\columnwidth]{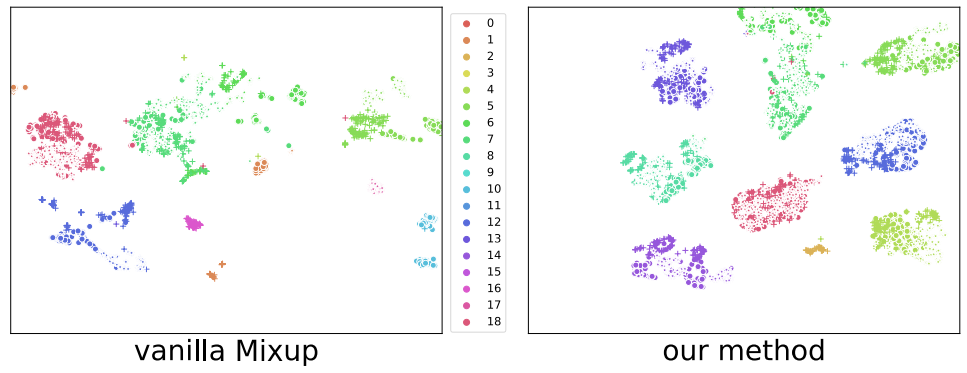}
}
\subfigure[Discrimination]{
\label{fig:viso}
\includegraphics[width=0.75\columnwidth]{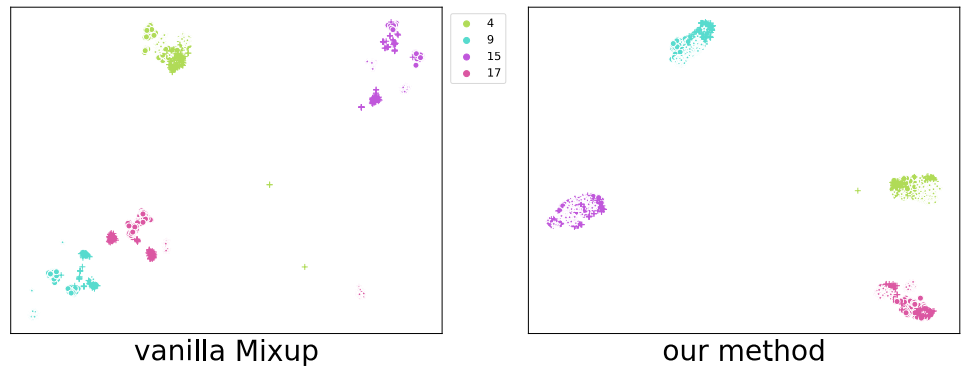}
}
\caption{Visualization of the t-SNE embeddings of DSADS dataset for target 3 in the Cross-Person setting. Each class is denoted by a color and each shape corresponds to a domain. \emph{Best viewed in color and zoom in.}}
\label{fig:vis}
\end{figure}



In \figurename~\ref{fig:vism}, different classes (points with different colors) have different semantic ranges, which demonstrates that semantic inconsistency exists in reality and we should use semantic-aware Mixup. From \figurename~\ref{fig:viso}, we can see that data points in color 17 with shape + are close to data points in color 9, which proves the existence of the second issue. And from the right figure, we can see our method mitigates this issue. 
\figurename~\ref{fig:vis} describes the phenomena and the improvements on the fourth task of DSADS in Cross-person setting also suggests that paying attention to semantic range and discrimination can bring better performance.\footnote{In contrast, paying no attention to semantic range and discrimination may suffer from some problems and have worse performance.}


\subsection{Ablation Study}

\begin{figure}[htbp]
\centering
\subfigure[SHAR]{
\includegraphics[width=0.31\textwidth]{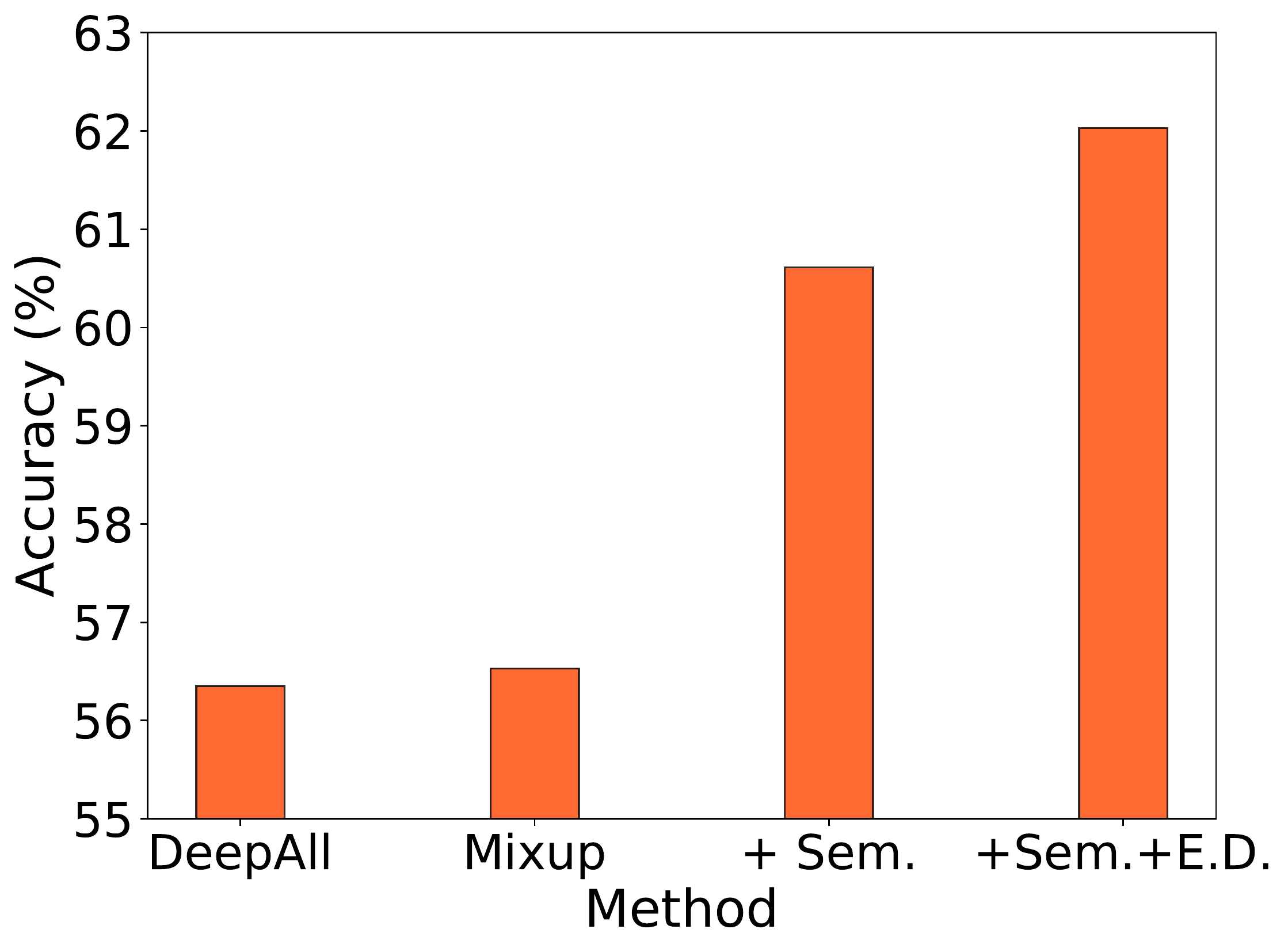}
\label{fig:ablat-sh}
}
\subfigure[DSADS]{
\includegraphics[width=0.31\textwidth]{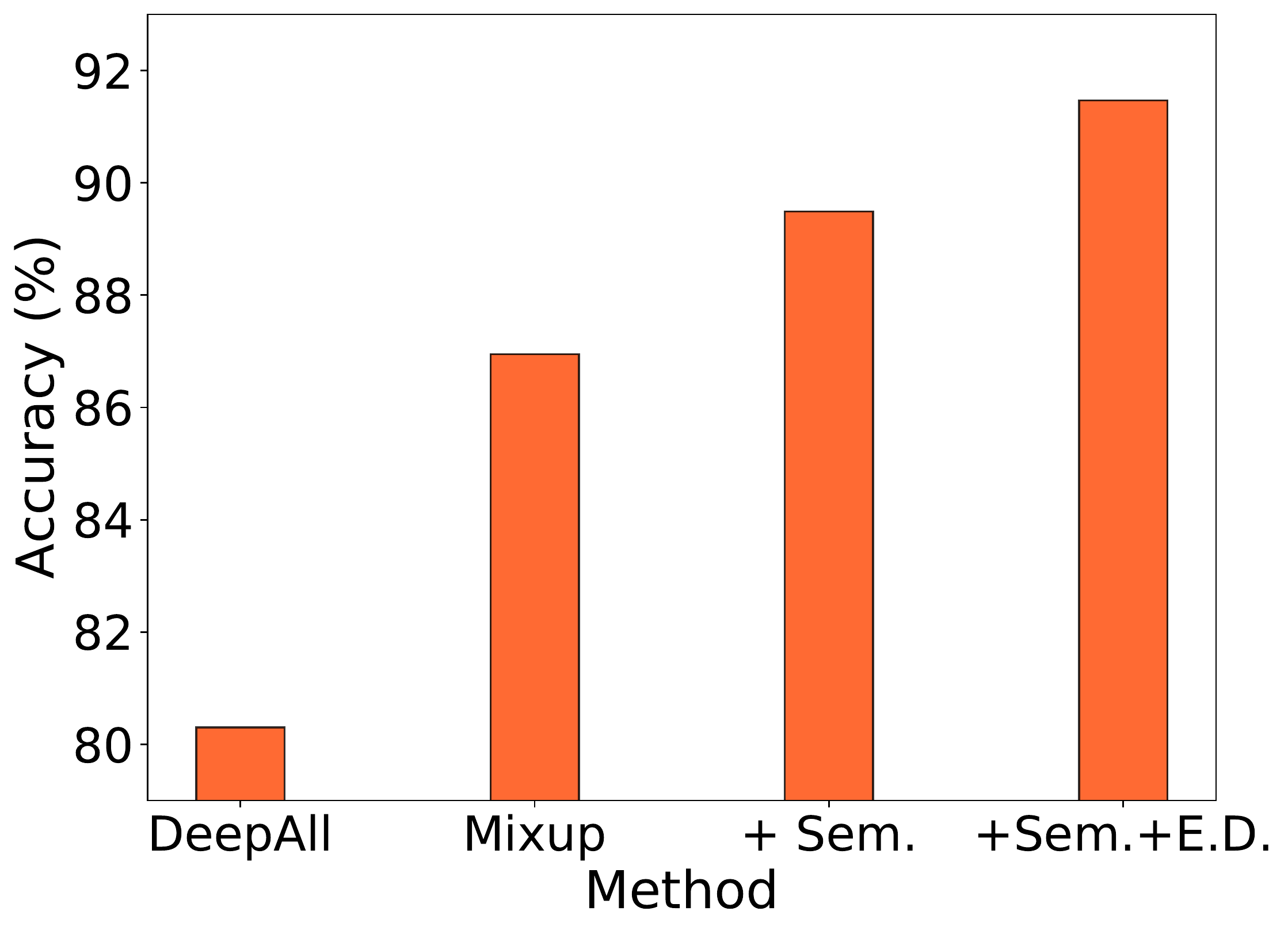}
\label{fig:ablat-ds}
}
\subfigure[PAMAP2]{
\includegraphics[width=0.31\textwidth]{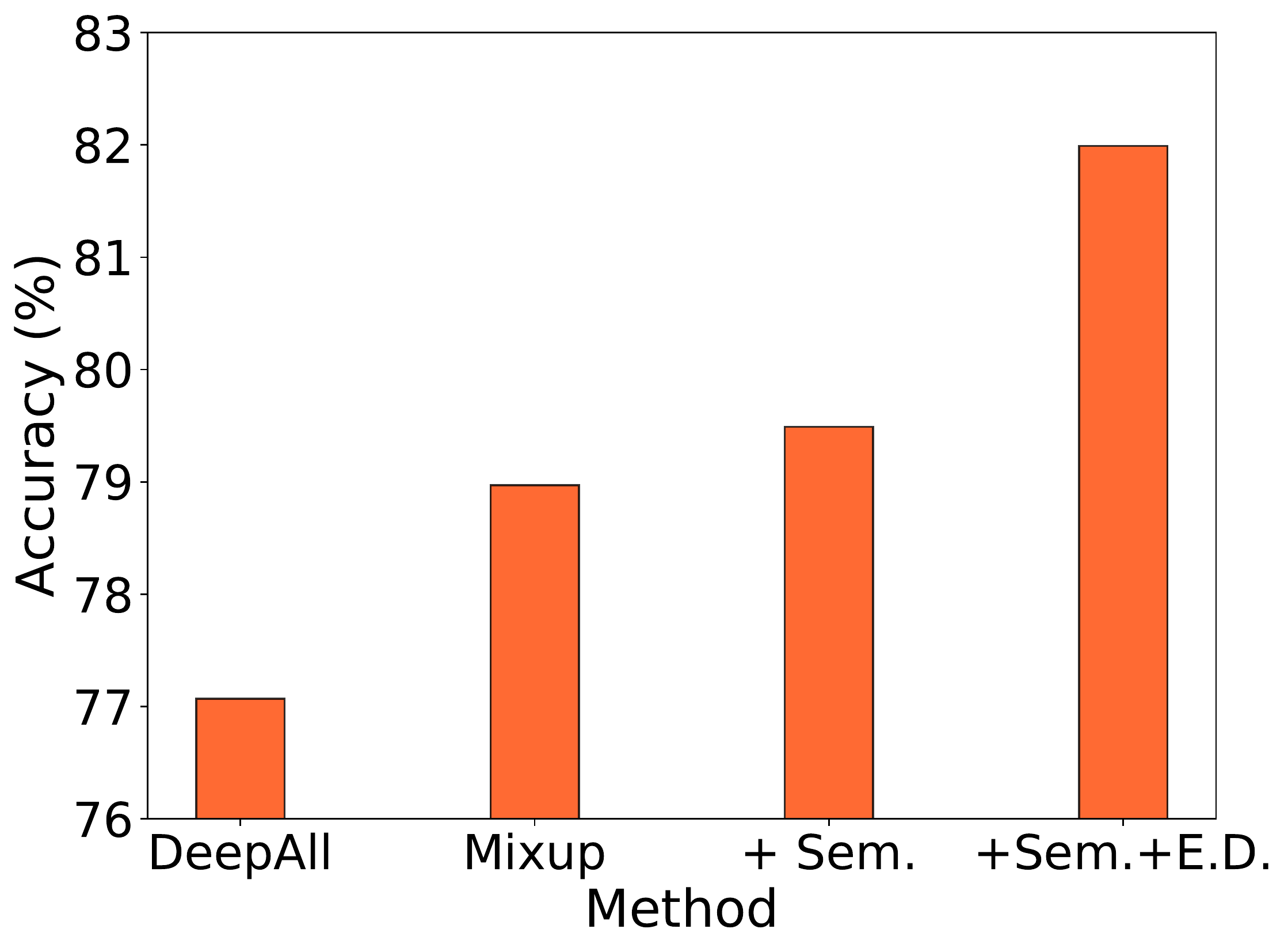}
\label{fig:ablat-pa}
}
\subfigure[USC-HAD]{
\includegraphics[width=0.31\textwidth]{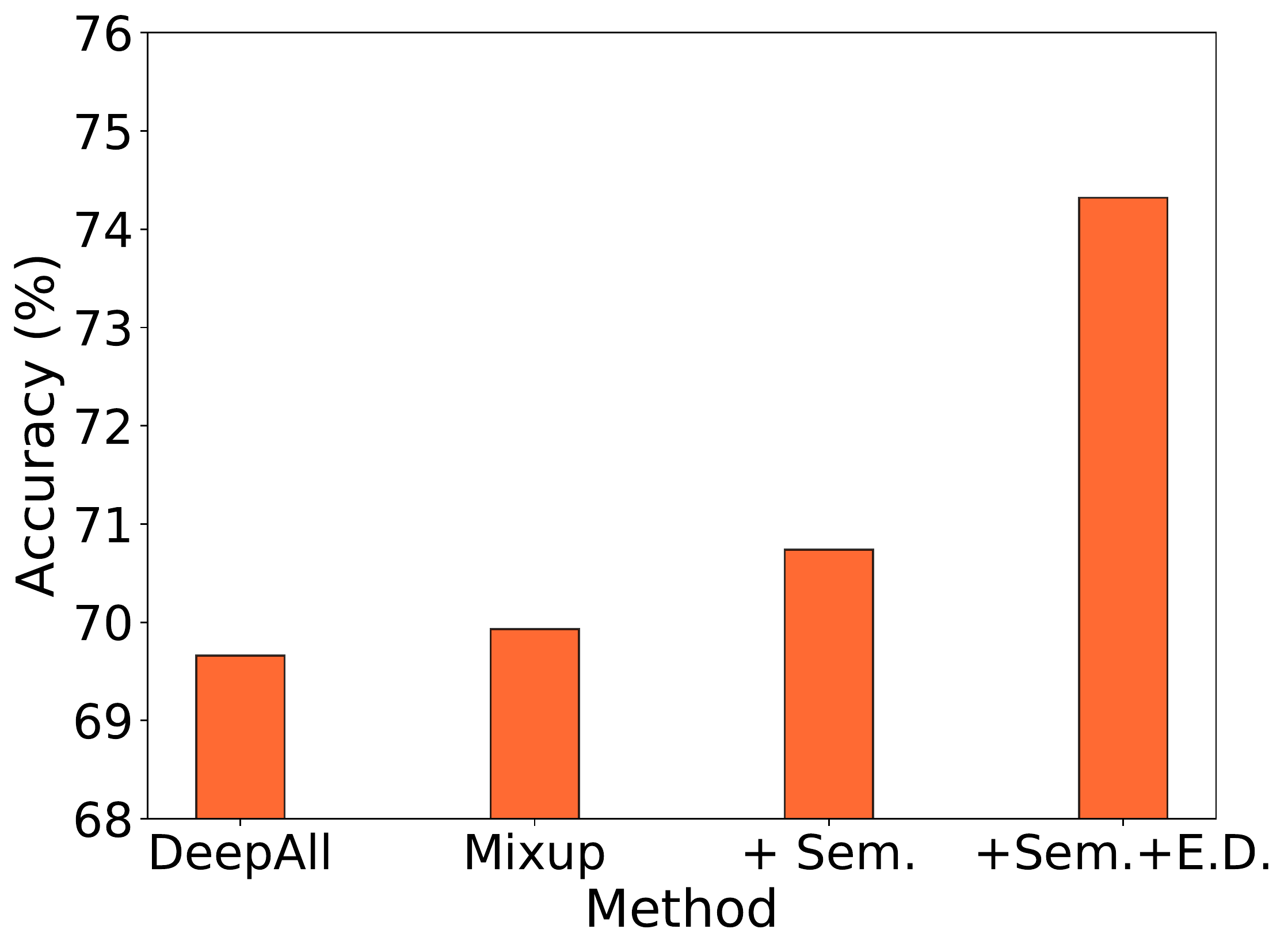}
\label{fig:ablat-usc}
}
\subfigure[Cross-Dataset]{
\includegraphics[width=0.31\textwidth]{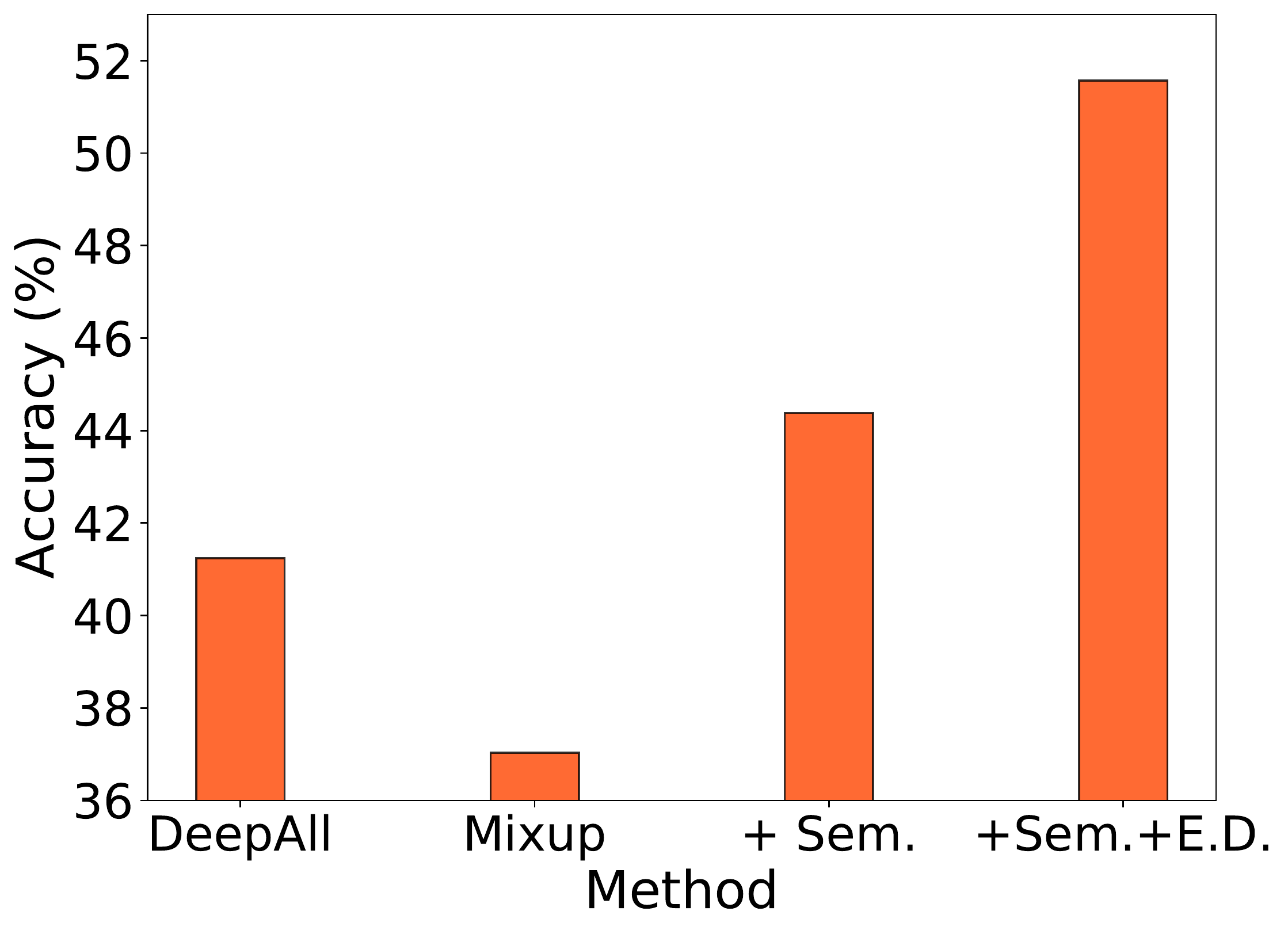}
\label{fig:ablat-cd}
}
\subfigure[Cross-Position]{
\includegraphics[width=0.31\textwidth]{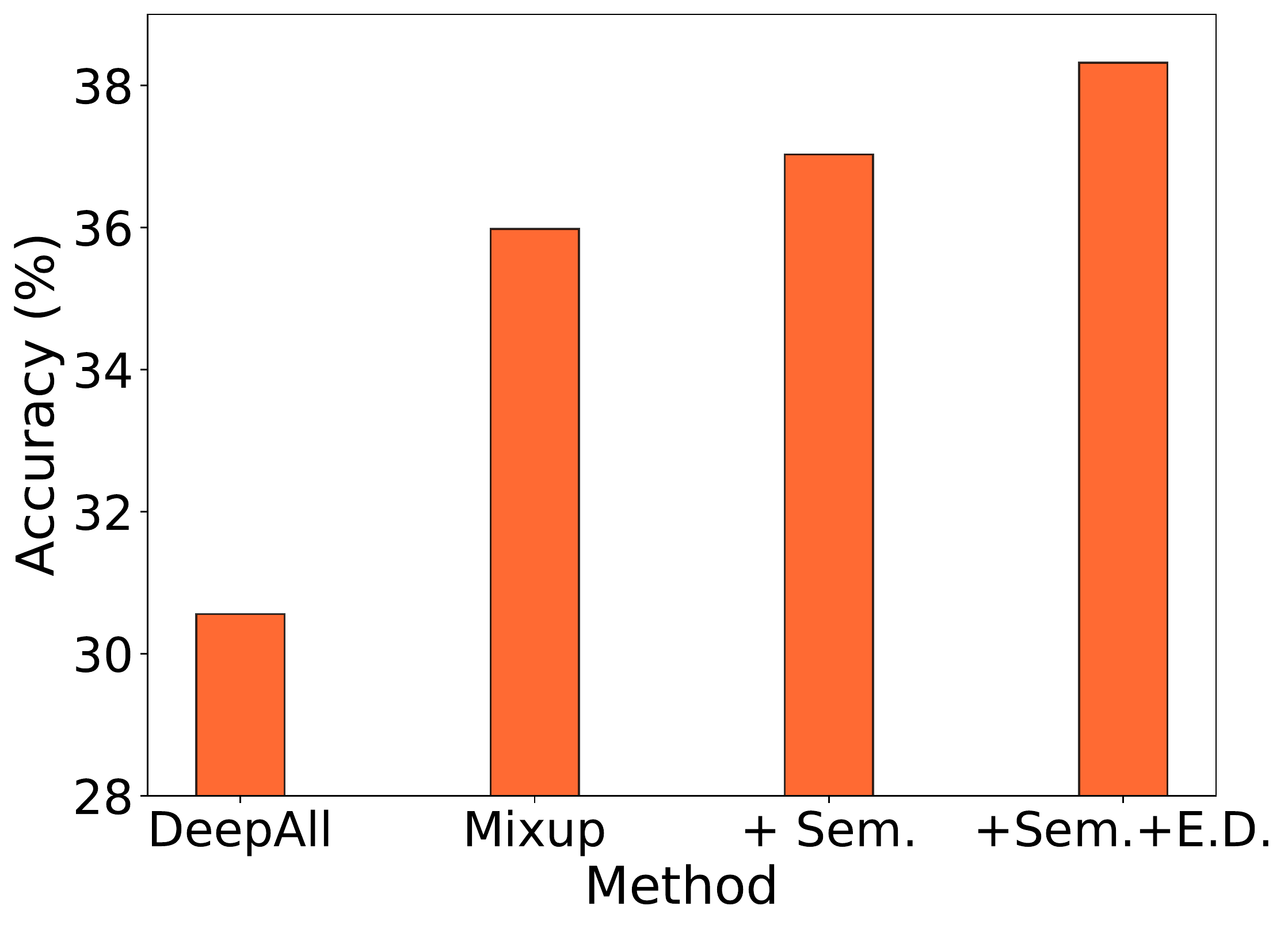}
\label{fig:ablat-cp}
}
\caption{Ablation study. \figurename~\ref{fig:ablat-sh}-\ref{fig:ablat-usc} are in the Cross-Person setting. \figurename~\ref{fig:ablat-cd} is in the Cross-Dataset setting while \figurename~\ref{fig:ablat-cp} is in the Cross-Position setting. `Sem.' and `E.D.' denotes semantic and enhancing discrimination, respectively.}
\label{fig:ablat}
\end{figure}

We perform ablation study in \figurename~\ref{fig:ablat}. 
\figurename~\ref{fig:ablat-sh} shows Mixup has a better performance than DeepAll while our Semantic-aware Mixup further improves accuracy on SHAR in the Cross-Person setting.
Moreover, with large margin loss to enhance discrimination, there exists another improvement compared with semantic-aware Mixup, which demonstrates that our method achieves the best average accuracy and the two components of our \method are both effective.
Results of ablation study on DSADS, PAMAP2, and USC-HAD in the Cross-Person setting and results in the Cross-Position setting (\figurename~\ref{fig:ablat-ds}, \figurename~\ref{fig:ablat-pa}, \figurename~\ref{fig:ablat-usc}, and \figurename~\ref{fig:ablat-cp} respectively) all demonstrate that increasing diversity of data can bring better performance and two components of our \method are both effective.
In \figurename~\ref{fig:ablat-cd}, Vanilla Mixup performs worse than DeepAll while ours achieve better results, which demonstrates that it is better to pay attention to semantic inconsistency and discrimination slackness when performing Mixup.

\subsection{Parameter Sensitivity Analysis}

\begin{figure}[htbp]
\centering
\subfigure[$\alpha$]{
\label{fig:sensalpha}
\includegraphics[width=0.23\columnwidth]{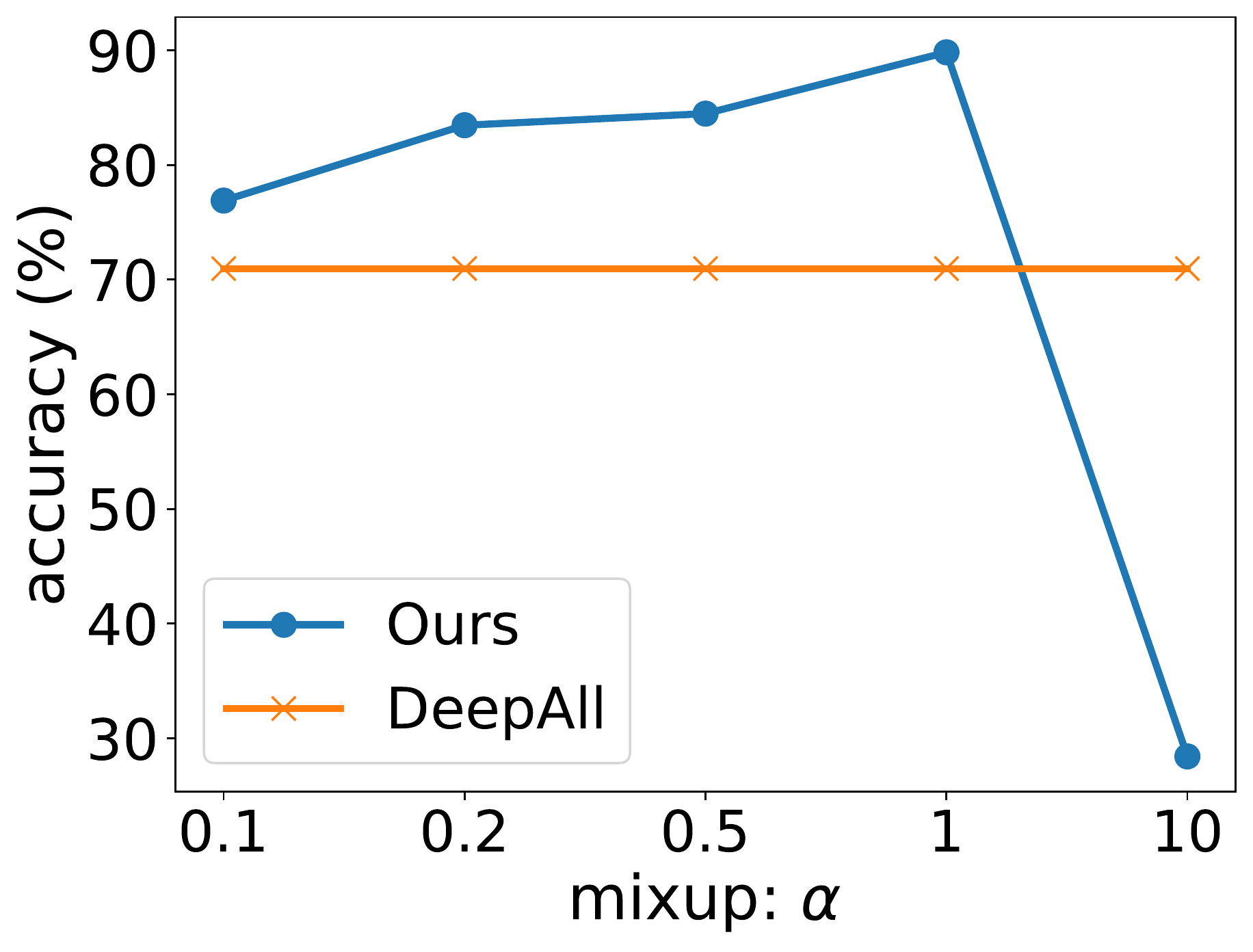}
}
\subfigure[Top c]{
\label{fig:sensk}
\includegraphics[width=0.24\columnwidth]{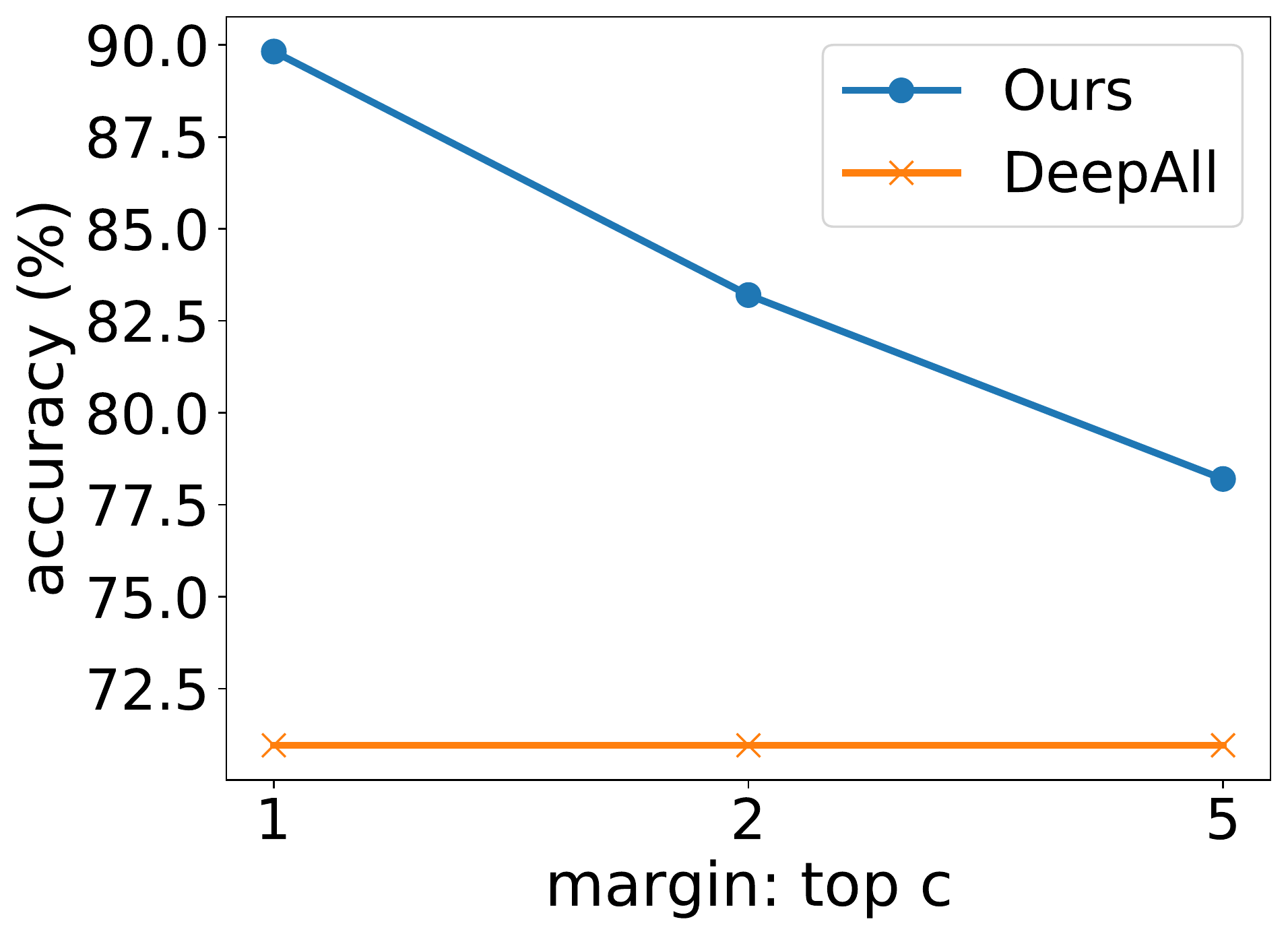}
}
\subfigure[$\gamma$]{
\label{fig:sensgamma}
\includegraphics[width=0.24\columnwidth]{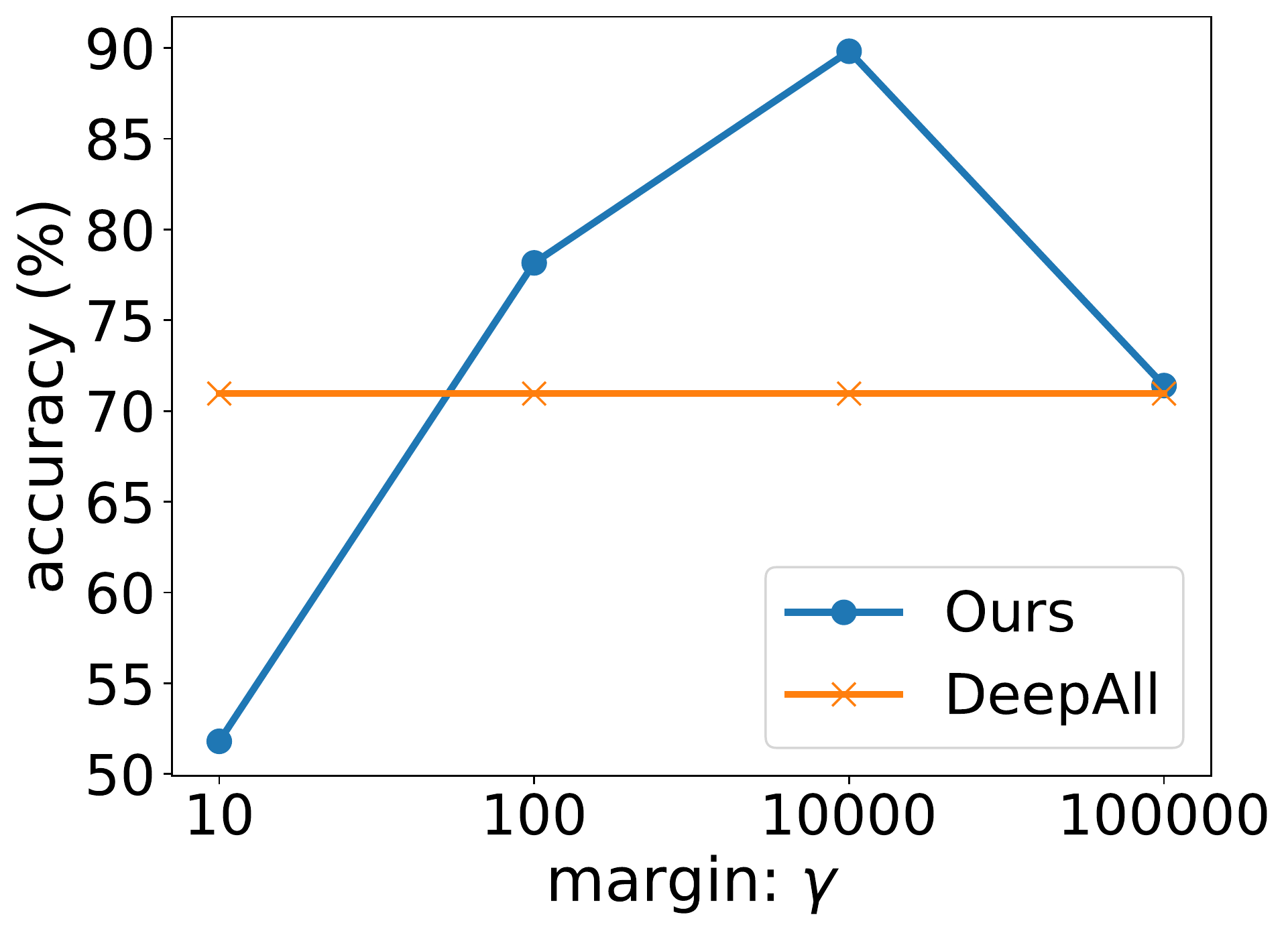}
}
\subfigure[Alternatives to compute $R_c$]{
\label{fig:sensradii}
\includegraphics[width=0.23\columnwidth]{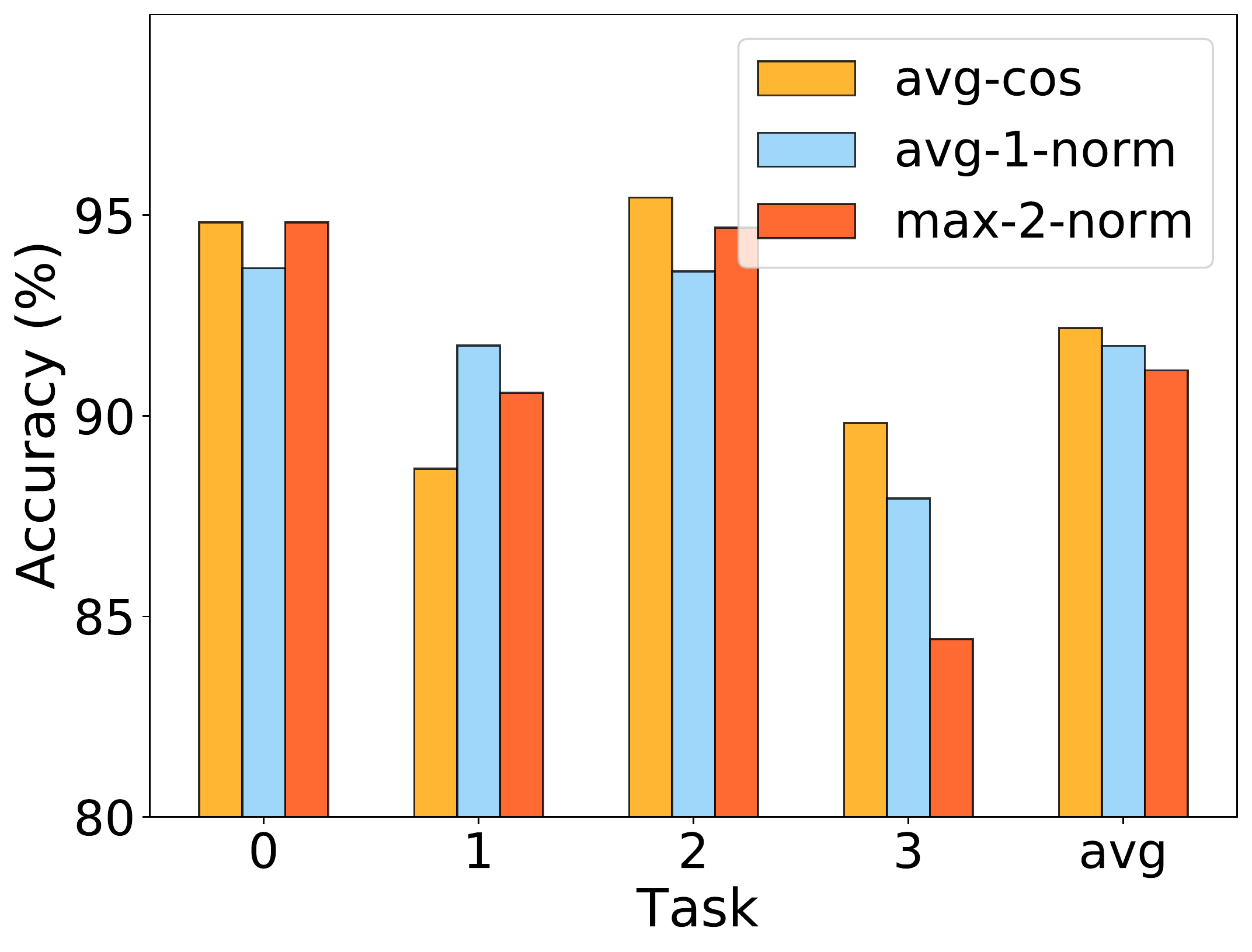}
}
\caption{Parameter sensitivity analysis (DSADS in the Cross-Person setting)}.
\label{fig:sens}
\end{figure}

We evaluate the parameter sensitivity of \method in \figurename~\ref{fig:sens}. 
There are mainly four hyperparameters in our method: 
the Beta distribution in Mixup ($\alpha$), the aggregation class number in \equationname~\eqref{eqa:approx} (top $c$), the required distance to boundaries in \equationname~\eqref{eqa:approx} ($\gamma$), and alternatives for computing activity semantic range in \equationname~\eqref{eqa:maxr} or \equationname~\eqref{eqa:avgr}.
From \figurename~\ref{fig:sensalpha}-\ref{fig:sensgamma}, we can see that the results with parameters around the highest points are all better than DeepAll. \figurename~\ref{fig:sensradii} demonstrates that different approaches of computing activity semantic ranges obtain different results and we should choose the right one for better results through validation. 
Please note that our method achieves the best performance no matter which approach of computing is selected.
In a nutshell, It demonstrates that \method is effective and robust that can be easily applied to real applications.

\subsection{Extensibility}
To demonstrate that our method is still a remark method in the common setting where training data and testing data share the same distribution, we conduct experiments on DSADS. 
We split DSADS into two parts, $20\%$ for testing and the rest for training.
When utilizing all rest data for training, the baseline accuracy is almost $100\%$.

\begin{figure}[htbp]
\centering
\includegraphics[width=0.35\textwidth]{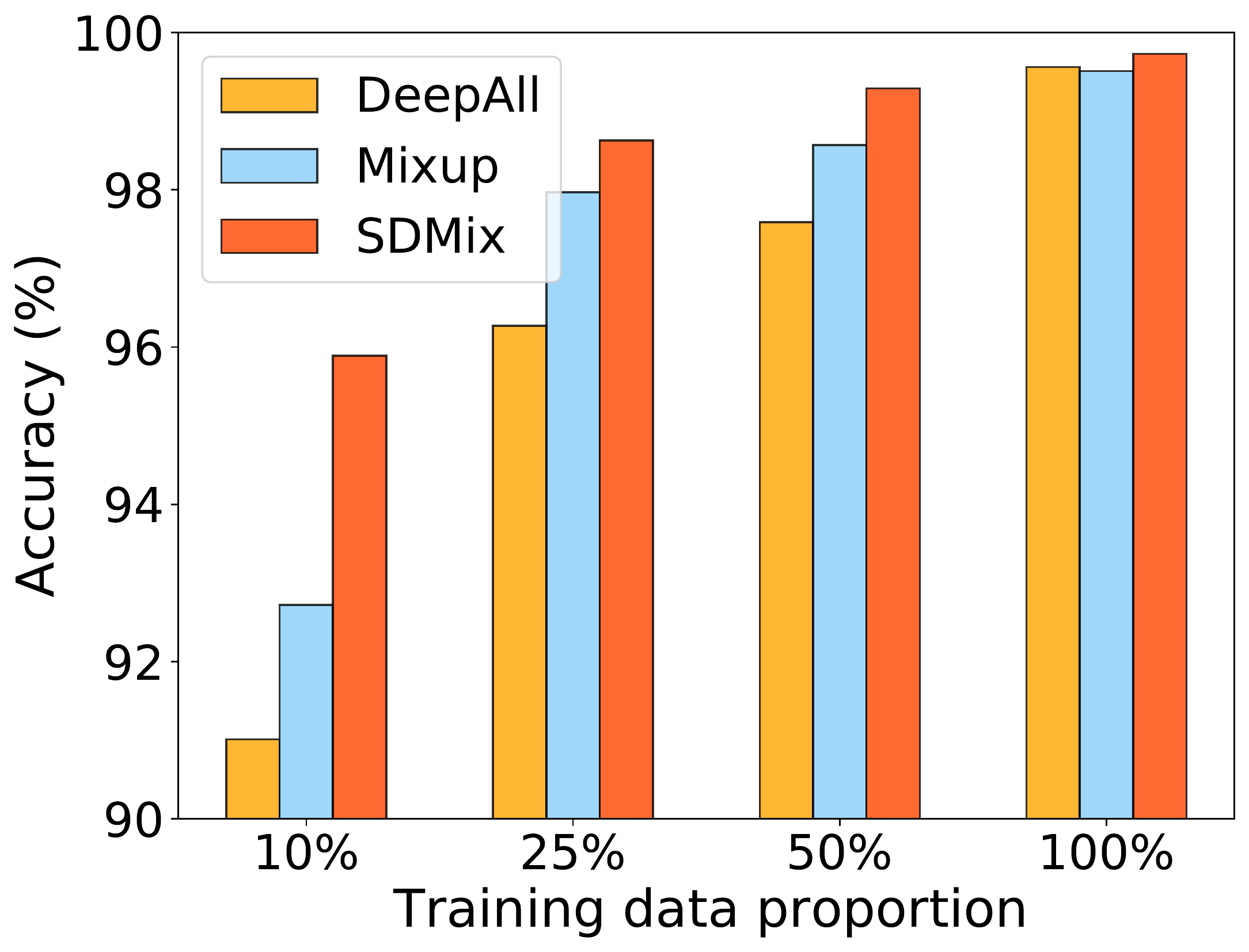}
\caption{Results on DSADS when training and testing data have the same distribution.}
\label{fig:common}
\end{figure}

To better show performance differences between our method and other state-of-the-art methods, we utilize $10\%, 25\%, 50\%, 100\%$ of rest data for training. 
To select the best model during training, $20\%$ of data serve as validation data while the rest data serve as the real training data. 
Therefore, when utilizing $10\%$ of rest data for training, only $6.4\%$ of all data are for training.
The results are shown in \figurename~\ref{fig:common} where we can see that our method achieves the best performance compared to ERM and Mixup. 
Our method has improvements with $4.88\%, 2.36\%, 1.7\%, 0.17\%$ compared to ERM while it has improvements with $3.17\%, 0.66\%, 0.72\%, 0.22\%$ compared to Mixup using $10\%, 25\%, 50\%, 100\%$ training data respectively.
As we can see, all methods achieve better performance with more training data used and our method achieves the best performance no matter how many training data are used. 
The results demonstrate that our method is still a remark method in a common setting.



\section{Conclusion}
\label{sec:concl}

In this paper, we proposed \method for generalizable sensor-based human activity recognition.
\method solves two critical challenges: semantic inconsistency and discriminative slackness.
Specifically, \method took the activity semantic range into consideration to learn more flexible interpolations.
Moreover, \method introduced the large margin loss to Mixup that enhances the discrimination.
Extensive experiments on cross-person, cross-dataset, and cross-position scenarios demonstrated the effectiveness of our method.
In the future, we plan to extend \method to more challenging HAR settings including cross-category and incremental learning.

\section*{ACKNOWLEDGMENTS}
This work is supported by the National Key Research and Development Plan of China No.2021YFC2501202, Natural Science Foundation of China (No. 61972383, No. 61902377, No. 61902379, No.62002187), Beijing Municipal Science \& Technology Commission No.Z211100002121171.

\bibliographystyle{ACM-Reference-Format}
\bibliography{imwut22}

\end{document}

%% file: sec_related.tex
\subsection{Human Activity Recognition}
Sensor-based HAR is receiving increasing attention over the years~\cite{wang2019deep}.
To alleviate the cost for data collection, there is growing interest in applying transfer learning and domain adaptation for HAR~\cite{cook2013transfer,chang2020systematic, wang2018deep}.
All existing domain adaptation and transfer learning-based HAR are assuming the availability of target domain data during training.
Very little attention is paid to generalizable HAR, i.e., when the test distribution is unavailable for training.
To our best knowledge, GILE~\cite{qian2021latent} is the first work for generalizable HAR that disentangled domain-agnostic and domain-specific features in the latent feature space. However, GILE is based on a variational auto-encoder which is more complicated in real applications. Our method does not use generative models and it can perform well on simple CNNs based on Mixup data augmentation.

\subsection{Domain Adaptation}
Domain adaptation has developed for many years and detailed surveys can be found in~\cite{pan2009survey,wilson2020survey}.
There is much prior work focusing on HAR with domain adaptation and a detailed survey can be found in~\cite{cook2013transfer}.
For cross-domain activity recognition, Wang et al.~\cite{wang2018stratified} proposed a stratified transfer learning (STL) algorithm that learns the class-wise feature transformation.
Wang et al.~\cite{wang2018deep} designed a unified framework for source domain selection and activity transfer.
Later, to select the most similar source domain to the target domain and perform accurately transfer activity, Qin et al.~\cite{qin2019cross} proposed an adaptive spatial-temporal transfer learning (ASTTL) approach.
Most recently, Lu et al.~\cite{lu2021cross} tried to make full use of data structures and matched substructures of data via optimal transport.
Although the above work can relieve distribution shifts, it cannot cope with situations when targets data is inaccessible. 

\subsection{Domain Generalization}
Domain generalization (DG) tries to learn a model from one or several different but related domains that will generalize well on unseen testing domains.
DG is different from leave-one-out-cross-validation (LOOCV)~\cite{wong2015performance} since LOOCV is a model selection technique which selects models via parts of training data while DG is to enhance generalization on unseen targets. 
Existing domain generalization methods can be grouped into three categories~\cite{wang2021generalizing}: data manipulation, representation learning, and learning strategy. 
Data manipulation contains two kinds of popular techniques: data augmentation~\cite{nazari2020domain,garg2021learn} which is mainly based on augmentation, randomization, transformation of input data, and data generation~\cite{qiao2020learning,liu2018unified} which generates diverse samples to help generalization. 
Representation learning also includes two kinds of representative techniques: domain-invariant representation learning~\cite{rahman2020correlation,li2018deep} which performs kernel, adversarial training, explicitly feature alignment between domains, or invariant risk minimization to learn domain-invariant representations and feature disentanglement~\cite{xu2014exploiting,ilse2020diva} which tries to disentangle the features into domain-shared or domain-specific parts for better generalization. 
Learning strategy mainly has three kinds of methods: ensemble learning~\cite{mancini2018best} which relies on the power of ensemble to learn a unified and generalized predictive function, meta-learning~\cite{wang2020meta,balaji2018metareg} which is based on the learning-to-learn mechanism to learn general knowledge by constructing meta-learning tasks to simulate domain shift, and gradient operation~\cite{huang2020self,shi2021gradient} which tries to learn generalized representations by directly operating on gradients. 
For more details, please refer to~\cite{wang2021generalizing}.
Despite the popularity of DG in recent years, most of the methods are proposed for computer vision tasks that do not consider the application of HAR.

\subsection{Mixup}
Mixup~\cite{zhang2018mixup} is a simple but effective technique for data augmentation. 
It extends the training distribution by incorporating the prior knowledge that linear interpolations of feature vectors should lead to linear interpolations of the associated targets. 
Although vanilla Mixup has shown its ability for domain generalization, many adapted versions are proposed for better performances.
Recent work~\cite{xu2020adversarial,wu2020dual} used the vanilla Mixup for domain adaptation without modifications.
Wang et al.~\shortcite{wang2020heterogeneous} mixed up samples across multiple source domains with two different sampling strategies to improve the generalization performance across different tasks. 
They perform well on computer vision while there is no Mixup method designed specifically for HAR.

%% file: tb_allresults.tex
\begin{table*}[t!]
\centering
\caption{Classification accuracy ($\pm$ std. error) for \emph{cross-person} HAR. The \textbf{bold} and \underline{underline} items denote the best and second-best results, respectively.}
\resizebox{\textwidth}{!}{%
\begin{tabular}{cccccccccccc}
\toprule
\multicolumn{1}{l}{} & Src & Tar & DeepAll & DANN & CORAL & ANDMask & GroupDRO & RSC & Mixup & GILE & \method \\ \midrule
\multirow{5}{*}{\rotatebox{90}{SHAR}} &1,2,3&0&57.98$\pm${\scriptsize 5.90}&55.99$\pm${\scriptsize 1.82}&57.90$\pm${\scriptsize 1.65}&58.33$\pm${\scriptsize 5.21}&54.60$\pm${\scriptsize 1.74}&57.81$\pm${\scriptsize 3.12}&\underline{58.68}$\pm${\scriptsize 1.91}&49.57$\pm${\scriptsize 3.57}&\textbf{65.80}$\pm${\scriptsize 0.70}\\
&0,2,3&1&57.29$\pm${\scriptsize 3.26}&50.20$\pm${\scriptsize 3.54}&\textbf{59.92}$\pm${\scriptsize 1.60}&48.89$\pm${\scriptsize 2.75}&58.72$\pm${\scriptsize 5.20}&55.06$\pm${\scriptsize 6.17}&55.69$\pm${\scriptsize 3.89}&48.02$\pm${\scriptsize 4.02}&\underline{59.86}$\pm${\scriptsize 2.40}\\
&0,1,3&2&\underline{69.96}$\pm${\scriptsize 2.20}&63.92$\pm${\scriptsize 3.72}&69.41$\pm${\scriptsize 1.65}&66.34$\pm${\scriptsize 2.85}&68.09$\pm${\scriptsize 2.63}&67.87$\pm${\scriptsize 2.08}&\underline{69.96}$\pm${\scriptsize 4.17}&65.44$\pm${\scriptsize 5.44}&\textbf{74.45}$\pm${\scriptsize 0.77}\\
&0,1,2&3&40.16$\pm${\scriptsize 1.23}&42.06$\pm${\scriptsize 2.46}&40.60$\pm${\scriptsize 0.33}&\underline{42.39}$\pm${\scriptsize 3.13}&42.17$\pm${\scriptsize 0.56}&40.27$\pm${\scriptsize 1.34}&41.83$\pm${\scriptsize 1.23}&39.27$\pm${\scriptsize 3.27}&\textbf{47.99}$\pm${\scriptsize 0.67}\\
&AVG&-&56.35$\pm${\scriptsize 1.78}&53.04$\pm${\scriptsize 2.16}&\underline{56.96}$\pm${\scriptsize 0.57}&53.99$\pm${\scriptsize 1.93}&55.89$\pm${\scriptsize 1.33}&55.25$\pm${\scriptsize 3.17}&56.53$\pm${\scriptsize 2.21}&50.54$\pm${\scriptsize 3.04}&\textbf{62.03}$\pm${\scriptsize 0.47} \\ \midrule

\multirow{5}{*}{\rotatebox{90}{DSADS}} &1,2,3&0&83.26$\pm${\scriptsize 4.22}&88.09$\pm${\scriptsize 4.84}&90.51$\pm${\scriptsize 2.00}&85.17$\pm${\scriptsize 1.05}&\underline{91.77}$\pm${\scriptsize 1.16}&84.21$\pm${\scriptsize 1.93}&88.08$\pm${\scriptsize 2.12}&79.67$\pm${\scriptsize 1.67}&\textbf{94.20}$\pm${\scriptsize 1.04}\\
&0,2,3&1&77.79$\pm${\scriptsize 0.95}&79.37$\pm${\scriptsize 2.57}&83.58$\pm${\scriptsize 5.20}&77.73$\pm${\scriptsize 1.94}&\underline{84.30}$\pm${\scriptsize 1.58}&78.79$\pm${\scriptsize 2.47}&80.95$\pm${\scriptsize 1.65}&75.00$\pm${\scriptsize 0.00}&\textbf{91.07}$\pm${\scriptsize 0.54}\\
&0,1,3&2&84.68$\pm${\scriptsize 3.01}&82.48$\pm${\scriptsize 3.14}&83.04$\pm${\scriptsize 4.84}&83.32$\pm${\scriptsize 5.38}&82.06$\pm${\scriptsize 8.51}&81.48$\pm${\scriptsize 4.46}&\underline{88.00}$\pm${\scriptsize 2.52}&77.00$\pm${\scriptsize 1.00}&\textbf{93.61}$\pm${\scriptsize 0.80}\\
&0,1,2&3&74.74$\pm${\scriptsize 3.78}&76.05$\pm${\scriptsize 6.18}&75.45$\pm${\scriptsize 1.37}&78.17$\pm${\scriptsize 3.87}&78.48$\pm${\scriptsize 0.23}&77.12$\pm${\scriptsize 1.94}&\underline{84.20}$\pm${\scriptsize 1.96}&67.00$\pm${\scriptsize 1.00}&\textbf{87.00}$\pm${\scriptsize 1.39}\\
&AVG&-&80.12$\pm${\scriptsize 1.08}&81.50$\pm${\scriptsize 2.84}&83.15$\pm${\scriptsize 1.32}&81.10$\pm${\scriptsize 3.03}&84.15$\pm${\scriptsize 2.15}&80.40$\pm${\scriptsize 1.88}&\underline{85.31}$\pm${\scriptsize 1.76}&74.65$\pm${\scriptsize 0.15}&\textbf{91.47}$\pm${\scriptsize 0.27} \\ \midrule

 \multirow{5}{*}{\rotatebox{90}{PAMAP2}} &1,2,3&0&\underline{89.02}$\pm${\scriptsize 0.60}&85.78$\pm${\scriptsize 3.60}&82.82$\pm${\scriptsize 6.85}&87.54$\pm${\scriptsize 0.80}&83.42$\pm${\scriptsize 2.08}&87.14$\pm${\scriptsize 2.29}&80.24$\pm${\scriptsize 10.41}&83.33$\pm${\scriptsize 0.33}&\textbf{89.14}$\pm${\scriptsize 0.51}\\
&0,2,3&1&75.80$\pm${\scriptsize 1.84}&75.86$\pm${\scriptsize 1.61}&77.43$\pm${\scriptsize 1.44}&77.97$\pm${\scriptsize 1.53}&74.95$\pm${\scriptsize 2.37}&77.31$\pm${\scriptsize 5.95}&\underline{78.77}$\pm${\scriptsize 3.81}&68.67$\pm${\scriptsize 0.67}&\textbf{84.99}$\pm${\scriptsize 0.29}\\
&0,1,3&2&51.83$\pm${\scriptsize 6.77}&49.47$\pm${\scriptsize 10.89}&48.14$\pm${\scriptsize 7.20}&45.36$\pm${\scriptsize 7.94}&50.59$\pm${\scriptsize 8.20}&55.11$\pm${\scriptsize 3.40}&\underline{55.53}$\pm${\scriptsize 4.37}&44.00$\pm${\scriptsize 2.00}&\textbf{63.11}$\pm${\scriptsize 2.27}\\
&0,1,2&3&82.98$\pm${\scriptsize 3.06}&84.64$\pm${\scriptsize 3.41}&84.47$\pm${\scriptsize 3.62}&84.01$\pm${\scriptsize 3.45}&82.66$\pm${\scriptsize 1.97}&83.64$\pm${\scriptsize 5.30}&\underline{86.63}$\pm${\scriptsize 0.68}&76.67$\pm${\scriptsize 0.67}&\textbf{88.77}$\pm${\scriptsize 0.22}\\
&AVG&-&74.91$\pm${\scriptsize 1.56}&73.94$\pm${\scriptsize 2.44}&73.21$\pm${\scriptsize 3.18}&73.72$\pm${\scriptsize 1.12}&72.90$\pm${\scriptsize 2.15}&\underline{75.80}$\pm${\scriptsize 2.67}&75.29$\pm${\scriptsize 3.28}&68.25$\pm${\scriptsize 1.00}&\textbf{81.50}$\pm${\scriptsize 0.53} \\ \midrule

\multirow{5}{*}{\rotatebox{90}{USC-HAD}} &1,2,3&0&79.90$\pm${\scriptsize 3.80}&80.54$\pm${\scriptsize 1.39}&79.02$\pm${\scriptsize 3.11}&79.46$\pm${\scriptsize 0.94}&\underline{81.57}$\pm${\scriptsize 1.45}&80.05$\pm${\scriptsize 2.57}&79.70$\pm${\scriptsize 2.66}&78.67$\pm${\scriptsize 0.67}&\textbf{84.09}$\pm${\scriptsize 0.66}\\
&0,2,3&1&59.04$\pm${\scriptsize 2.50}&61.36$\pm${\scriptsize 3.48}&59.87$\pm${\scriptsize 3.89}&60.01$\pm${\scriptsize 4.69}&58.96$\pm${\scriptsize 3.45}&57.04$\pm${\scriptsize 2.60}&57.43$\pm${\scriptsize 4.14}&\textbf{63.00}$\pm${\scriptsize 1.00}&\underline{62.86}$\pm${\scriptsize 0.74}\\
&0,1,3&2&72.57$\pm${\scriptsize 1.27}&75.90$\pm${\scriptsize 0.45}&74.41$\pm${\scriptsize 1.64}&73.74$\pm${\scriptsize 1.19}&71.59$\pm${\scriptsize 1.61}&74.25$\pm${\scriptsize 0.86}&73.83$\pm${\scriptsize 1.19}&\textbf{77.00}$\pm${\scriptsize 1.00}&\underline{76.46}$\pm${\scriptsize 0.97}\\
&0,1,2&3&67.15$\pm${\scriptsize 1.29}&\underline{69.33}$\pm${\scriptsize 4.06}&60.42$\pm${\scriptsize 6.70}&65.58$\pm${\scriptsize 0.54}&59.47$\pm${\scriptsize 2.44}&66.99$\pm${\scriptsize 1.86}&59.91$\pm${\scriptsize 2.98}&61.67$\pm${\scriptsize 0.67}&\textbf{71.90}$\pm${\scriptsize 0.05}\\
&AVG&-&69.67$\pm${\scriptsize 1.38}&\underline{71.78}$\pm${\scriptsize 1.97}&68.43$\pm${\scriptsize 1.81}&69.70$\pm${\scriptsize 1.02}&67.90$\pm${\scriptsize 0.51}&69.58$\pm${\scriptsize 0.19}&67.72$\pm${\scriptsize 1.60}&70.08$\pm${\scriptsize 0.08}&\textbf{73.83}$\pm${\scriptsize 0.45} \\ \midrule
 \multicolumn{1}{l}{} & AVG all & - & 70.26&70.07&70.44&69.63&70.21&70.26&\underline{71.21}&65.88&\textbf{77.21} \\ \bottomrule
 \end{tabular}}%
\label{tb-crossperson}
\end{table*}

%% file: tb_crossdataset.tex
\begin{table*}[t!]
\centering
\caption{Classification accuracy for \emph{cross-dataset} HAR. The \textbf{bold} and \underline{underline} items are the best and second-best results.}
\resizebox{\textwidth}{!}{%
\begin{tabular}{cccccccccc}
\toprule
Source            & Target & DeepAll            & DANN  & CORAL & ANDMask & GroupDRO    & RSC         & Mixup       & SDMix          \\ \midrule
DSADS, USC, PAMAP2 & UCI-HAR    & \underline{46.06} & 39.10  & 44.44 & 43.22   & 33.20       & 45.28 & 40.24       & \textbf{46.41}          \\ 
USC, UCI-HAR, PAMAP2   & DSADS  & 29.73          & 39.46 & 26.35 & 41.66   & \underline{51.41} & 33.10       & 37.35       & \textbf{52.66} \\
DSADS, USC, UCI-HAR   & PAMAP2  & 43.84          & 36.61 & 32.93 & 40.17   & 33.80       & \underline{45.94} & 23.12       & \textbf{53.65} \\ 
DSADS, UCI-HAR, PAMAP2 & USC    & 45.33          & 41.82 & 29.58 & 33.83   & 36.74       & 39.70       & \underline{47.39} & \textbf{53.54} \\ \midrule
AVG                  & -    & \underline{41.24} & 39.25 & 33.32 & 39.72   & 38.79       & 41.01       & 37.03       & \textbf{51.57} \\
 \bottomrule
\end{tabular}%
}
\label{tb-crossdataset}
\end{table*}

%% file: tb_crossposition.tex
\begin{table*}[t!]
\centering
\caption{Classification accuracy for \emph{cross-position} HAR. The \textbf{bold} and \underline{underline} items are the best and second-best results.}
\resizebox{.9\textwidth}{!}{%
\begin{tabular}{cccccccccc}
\toprule
Source  & Target & DeepALL & DANN  & CORAL & ANDMask     & GroupDRO & RSC   & Mixup          & SDMix          \\ \midrule
1,2,3,4 & 0      & 41.52   & 45.45 & 33.22 & \underline{47.51} & 27.12    & 46.56 & \textbf{48.77} & 47.50           \\
0,2,3,4 & 1      & 26.73   & 25.36 & 25.18 & 31.06       & 26.66    & 27.37 & \underline{34.19}    & \textbf{36.10}  \\
0,1,3,4 & 2      & 35.81   & 38.06 & 25.81 & \underline{39.17} & 24.34    & 35.93 & 37.49          & \textbf{42.53} \\
0,1,2,4 & 3      & 21.45   & 28.89 & 22.32 & \underline{30.22} & 18.39    & 27.04 & 29.50           & \textbf{34.52} \\
0,1,2,3 & 4      & 27.28   & 25.05 & 20.64 & 29.90        & 24.82    & 29.82 & \underline{29.95}    & \textbf{30.93} \\ \midrule
AVG     & -      & 30.56   & 32.56 & 25.43 & 35.57       & 24.27    & 33.34 & \underline{35.98}    & \textbf{38.32} \\
 \bottomrule
\end{tabular}%
}
\label{tb-crossposition}
\end{table*}